\documentclass[nonatbib]{article}

\usepackage[final]{neurips_2024}

\usepackage[utf8]{inputenc} 
\usepackage[T1]{fontenc}    
\usepackage{hyperref}       
\usepackage{url}            
\usepackage{booktabs}       
\usepackage{amsfonts}       
\usepackage{nicefrac}       
\usepackage{microtype}      
\usepackage{xcolor}         
\usepackage{stmaryrd}       
\usepackage{placeins}       

\usepackage{amsmath}        
\usepackage{amssymb}        
\usepackage{graphicx}       
\usepackage[super]{nth}     

\DeclareMathOperator*{\zargmax}{argmax}

\newcommand{\zpartderivw}[1]{\frac{\partial}{\partial \textbf{w}}{#1}}
\newcommand{\zpartderiva}[1]{\frac{\partial {#1}}{\partial{a_i}}}
\newcommand{\concat}[1]{||#1||}
\newcommand{\softmax}[1]{s(#1)}
\newcommand{\softmaxb}[1]{s[#1]}
\newcommand{\relu}[1]{relu(#1)}
\newcommand{\relub}[1]{relu\biggl[#1\biggr]}

\newcommand{\zres}[2]{\Phi^{#1}_{Res}(#2;\Omega)}
\newcommand{\zdens}[2]{\Phi^{#1}_{Dens}(#2;\Omega)}
\newcommand{\zdenst}[2]{\Phi^{#1}_{DensTran}(#2;\Omega)}
\newcommand{\zact}[2]{\textbf{#2}^{(#1)}}
\newcommand{\zconv}[2]{a^{(#1)}(#2)}
\newcommand{\zli}[1]{#1\biggl( 1 +\softmaxb{#1} \biggr)}
\newcommand{\zcount}[1]{\mathcal{N}(#1)}
\newcommand{\iverson}[1]{\llbracket#1\rrbracket}
\newcommand{\energy}{\mathbb{E}}
\newcommand{\explained}[1]{\overset{\therefore}{#1}}
\newcommand{\causeffect}[2]{m^{(#1)}_#2(C^{(#1-1)}_#2)}
\newcommand{\causality}[2]{\widehat{m}^{(#1)}_#2(C^{(#1-1)}_#2)}
\newcommand{\fact}[2]{m^{(#1)}_#2(C^{(#1-1)}_#2)}
\newcommand{\argmax}[2]{\zargmax_#1[#2]}
\newcommand{\argsort}[1]{arg\,sort[#1]}
\newcommand{\imagespace}[3]{\mathbb{R}^{#1\times#2\times#3}}
\newcommand{\kernelspace}[4]{\mathbb{R}^{#1\times#2\times#3\times#4}}
\newcommand{\matrixspace}[2]{\mathbb{R}^{#1\times#2}}
\newcommand{\matrixspacediscrete}[2]{\mathbb{N}^{#1\times#2}}
\newcommand{\zldfvector}[2]{\textbf{d}^{(#1)}_{#2}}

\newcommand{\zfmotifppm}[2]{\textbf{M}^{(#1)}_{PPM}(#2)}

\newcommand{\ci}[1]{\scriptsize{$\pm#1$}\normalsize}
\newcommand{\modelsize}[1]{#1\scriptsize MP \normalsize}
\newcommand{\convdescr}[5]{[#1\times#2\,/#3\,|#4 \to #5]}

\title{Learning local discrete features in explainable-by-design convolutional neural networks}

\author{%
  Pantelis I.~Kaplanoglou$^1$  \thanks{Primary contribution $^1$Department of Information and Electronic Engineering,
  International Hellenic University, Sindos, Greece
  }
  \quad
  Konstantinos~Diamantaras$^{1}$ \\
  \texttt{$^1$International Hellenic University}\\  
  \texttt{\{pikaplanoglou, kdiamant\}@ihu.gr}\\
  }

\begin{document}

\maketitle

\begin{abstract}
Our proposed framework attempts to break the trade-off between performance and explainability by introducing an explainable-by-design convolutional neural network (CNN) based on the lateral inhibition mechanism. 
The ExplaiNet model consists of the predictor, that is a high-accuracy CNN with residual or dense skip connections, and the explainer probabilistic graph that expresses the spatial interactions of the network neurons.
The value on each graph node is a local discrete feature (LDF) vector, a patch descriptor that represents the indices of antagonistic neurons ordered by the strength of their activations, which are learned with gradient descent. 
Using LDFs as sequences we can increase the conciseness of explanations by repurposing EXTREME, an EM-based sequence motif discovery method that is typically used in molecular biology. 
Having a discrete feature motif matrix for each one of intermediate image representations, instead of a continuous activation tensor, allows us to leverage the inherent explainability of Bayesian networks. By collecting observations and directly calculating probabilities, we can explain causal relationships between motifs of adjacent levels and attribute the model's output to global motifs.
Moreover, experiments on various tiny image benchmark datasets confirm that our predictor ensures the same level of performance as the baseline architecture for a given count of parameters and/or layers. 
Our novel method shows promise to exceed this performance while providing an additional stream of explanations. In the solved MNIST classification task, it reaches a comparable to the state-of-the-art performance for single models, using standard training setup and 0.75 million parameters.
\end{abstract}

\section{Introduction}
Deep learning has yielded high accuracy models, that was the initial prerequisite for the applicability of Artificial Intelligence (AI). With the advent of the transformer \cite{lin_survey_2022} architecture, the technical robustness of state-of-the-art models allows for real-world applications in all domains, including Computer Vision \cite{khan_transformers_2022}. Nevertheless, there are still challenges to be addressed for AI to become trustworthy \cite{li_trustworthy_2023}, towards social robustness. This is the focus of the evolving field of Explainable Machine Learning (ExML) that falls under the scientific branch of Explainable AI (XAI). Amongst others, we can identify seven “showstopper” issues for the seamless integration of AI into society; reliability, trustworthiness, bias, privacy,  physical security, human manipulation, and ethics. 
Their individual needs require explainability, that is a superset of interpretability which provides human understandable insight on models.
The main difference is completeness of explanations \cite{gilpin_explaining_2019}\cite{yeh_completeness-aware_2020} which can be defined as the capability of providing interpretations for any processing node of the model, from input to output or vice versa . 
To understand what an explainable model should do, we can draw equivalence to regular software, where the source code is inherently explainable. State-of-the-art models should offer “debugging” capabilities for defects caused by outlier or adversarial samples, overfitting, and \textit{confabulations} \cite{smith_hallucination_2023}, which for humans are attributed to overlearning. This capability will increase reliability, allow us to investigate dataset/model bias, and in turn ensure physical security.

The approach of explainable-by-design deploys inherently explainable models, for example a simple linear SVM classifier with its comprehensible decision surface. Yet, the performance of non-black box models follows a trend; increasing explainability decreases performance, which is known as the Performance-Explainability Trade-off (PET). Considering the carbon footprint of training models \cite{luccioni_estimating_2023}, a third dimension of time creates a new trade-off called PET+ \cite{crook_revisiting_2023}. It is already known that more training time leads to more performance and considering the extra effort for providing interpretations or explanations, more time is needed for more explainability. This creates a new requirement, to assess the model efficiency together with accuracy and explainability.

This work’s main contributions can be summarized in the following points:
\begin{itemize}
    \item We propose a novel lateral inhibition (LIN) layer that can be incorporated into the design of any CNN model. We prove that it regulates gradient descent learning of local discrete feature (LDF) vectors, that contain indices of neurons ordered by activation strength.
    \item We introduce ExplaiNet, an explainable-by-design classifier. The predictor is a CNN which provides increased accuracy through non-explainable continuous activations, that are synchronously discrete image representations used by an explainer probabilistic graph.
    \item We have experimented with various ExplaiNets of two different architectures in tiny image datasets, using various combinations of features per layer, and observed equal or greater prediction accuracy compared to the baseline CNN.
    \item We increase the conciseness of explanations by bringing the concept of sequence motifs from the field of Molecular Biology. These are learned by repurposing any existing unsupervised Expectation-Maximization algorithm for motif discovery.
    \item We evaluate the fidelity of discrete feature motifs (FMotifs) in all intermediate levels of the network, to provide causal explanations for the emergence of FMotifs and their contribution to the predicted class.
\end{itemize}

\setlength{\belowcaptionskip}{-12pt}
\begin{figure}[!b]
    \centerline{\includegraphics[width=12.3cm]{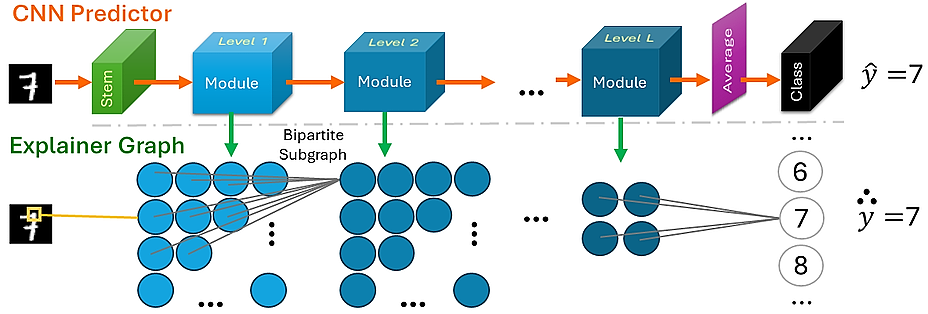}}
	\caption{Overview of the ExplaiNet model. A black-box feed forward (orange arrows) neural network predictor, that offers high prediction accuracy. Streams of discrete features (green arrows)  provide values to nodes of the probabilistic explainer graph that uses them to explain predictions and intermediate features. The nodes are mapped to spatial positions of the input at each level.}
	\label{fig:explainet}
\end{figure}

\section{Preliminaries}
\subsection{Terms and notation}
We present all terminology and notations used in the rest of the paper. The 2D convolution operation in CNNs is a linear function $\zconv{k}{\cdot}$ that moves a window over the input tensor $\textbf{A}^{(k)}$ of layer $k=1,…, K$, through positions $x$,$y$. The window corresponds to a tensor slice $\textbf{X}^{(k)}_{x,y}$ that is the \textit{receptive field} shared by all neurons operating at $x$,$y$, which form a \textit{hypercolumn}, the respective term in Neuroscience. The hypercolumn has a a continuous activation vector $\textbf{a}^{(k)}_{x,y}$ and the weights of the convolution operation are stored in the  kernel $\textbf{W}^{(k)} \in \kernelspace{n}{n}{c_{in}}{c_{out}}$. Considering activation values as features, the term for $\textbf{a}^{(k)}_{x,y}$ in Computer Vision is \textit{patch descriptor}. A CNN starts with the \textit{stem}, followed by \textit{blocks} of feature extraction, each block has a stack of \textit{neural modules}, that are composed of learnable and non-learnable layers. The network's representation output is fed to the classifier's fully connected neurons, that have the softmax activation function which is noted in this work as $\softmax{\cdot}$. 

\textbf{Residual convolutional module:} Without normalization, a residual network module \cite{he_deep_2016} is a non-linear transformation 
\footnotesize
$\zres{(l)}{\zact{k-2}{X}}=\relu{\zconv{k}{ \relu{ \zconv{k-1}{\zact{k-2}{X}} } } + \zact{k-2}{X} }$ 
\normalsize
where $l$ is the \textit{level} of the intermediate image representation, $\relu{\cdot}$ a rectifier function and $\Omega$ a set of weights and bias parameters, which are implicitly included in convolution functions
\footnotesize $\zconv{k}{\cdot}$. \normalsize

\textbf{Dense convolutional module:}
A densely connected block \cite{huang_densely_2017} has $\tau$ stacked modules
\footnotesize
$\zdens{(l)}{\zact{k-1}{X}}=\zconv{k}{\relu{\zact{k-1}{X}}}$. \normalsize
Their output is concatenated in 
\footnotesize
$\textbf{B}_{Dens}^{(k)}(\zact{l-\tau}{X})=\concat{\zdens{(k)}{\textbf{A}^{(k-1)}},\zdens{(k-1)}{\textbf{A}^{(k-2)}},..,\zdens{(k-\tau+1)}{\textbf{X}^{(k-\tau)}}}$.
\normalsize
Between subsequent blocks there are transition modules
\footnotesize
$\zdenst{k+1}{\textbf{B}_{Dens}^{(k)}}$
\normalsize
that additionally perform spatial downsampling.

\subsection{Gene sequence motif discovery}
In the field of Molecular Biology, patterns in genome sequences that are important to explain a biological function are called \textit{sequence motifs} \cite{dhaeseleer_what_2006}. These are recurring in genomic data, and its presence is used to investigate causality of some observable biological outcome, such as gene expression. Motifs are stochastic sequences, where each gene symbol has a probability of occurrence at a specific position of the motif. This is represented in $\textbf{M}_{PPM} \in \matrixspace{n_{motif}}{\delta}$, a Position-specific Probability Matrix (PPM) for a motif sequence length of $n_{motif}$ where each position can have one of $\delta$ possible discrete values; for DNA nucleotides \{A,T,C,G\} we have $\delta=4$. The PPM is used to generate a \textit{motif logo}, that visually depicts the probability of symbols at each position.

The common approach of motif discovery is unsupervised learning, with algorithms that are based on Expectation-Maximization (EM) \cite{bailey_unsupervised_1995}. The popular MEME algorithm \cite{bailey_meme_2006} has quadratic time complexity and several improvements were proposed, namely DREME \cite{bailey_dreme_2011}, EXTREME \cite{quang_extreme_2014} and STREME \cite{bailey_streme_2021}. The problem of EM's sensitivity to bootstrap conditions is alleviated by carefully choosing them from candidate sequences, what is known in this context as \textit{seeding}. The more recent EXTREME algorithm employs online EM \cite{cappe_-line_2009} that has linear time complexity.

\subsection{Metrics}

\textbf{Fidelity to Output:}
A criterion of how well a surrogate explainer pertains to the predictor classification output is fidelity to output (FTO), as suggested in \cite{parekh_framework_2021}. In an image dataset $D={\{\textbf{X}(i)\}}$, the predicted class for sample $i$ is $\hat{y}_i$ and $\explained{y}_i(l)$ is a class index that has been explained based on features of level $l$, where $\iverson{.}$ is the Iverson bracket. The fidelity to output (FCO) metric is:
\begin{equation}
\scriptsize
    \label{eq:1}
    fto^{(l)}(D) = \frac{\sum_{i}^{n}{ \iverson{\hat{y}_i=\explained{y}_i(l)} }  }{|D|}
\normalsize    
\end{equation}

\textbf{Fidelity of Cause to Effect:}
To express how well an observed feature in a level $\causeffect{l}{i}$, that is considered an effect of a model's function, can be attributed to a set of cause features $C^{(l-1)}_{i}={\{m^{(l-1)}_j\}}$ from the previous level, we introduce the fidelity of cause to effect metric (FCE):
\begin{equation}
\scriptsize
    \label{eq:2}
    fce^{(l)}(D) = \frac{1}{n_{fx}}\sum_{i}^{n_{fx}}{ \iverson{\fact{l}{i}=\causality{l}{i}} }
\normalsize       
\end{equation}
where $n_{fx}$ is the total count of unique effects observed in the dataset, $\fact{l}{i}$ the fact of an effect occurrence, and $\causality{l}{i}$ the explained effect that is inferred by a probabilistic explanation model from the same observed causes.

\textbf{Relative Model Efficiency:}
To evaluate parameters efficiency in a group of models G that have been trained on the same task, we introduce the relative model efficiency (RME) index. A model's metric $\mu_{i}$ is normalized with the minimum and maximum value in the group: 
$\nu_{i} =\frac{\mu_{i}-\mu_{max}}{\mu_{max}-\mu_{min}}$. The size of the model $s$ is in millions of parameters (\modelsize{}), and its magnitude is reduced in the denominator to avoid overpenalizing large but accurate models. For a model $i \in G$ the RME is:
\begin{equation}
\scriptsize
    \label{eq:3}
    rme(i, G) = \frac{\nu_i(2^{\nu_i} - 1)}{\sqrt{s}}
\normalsize     
\end{equation}


\section{Proposed Framework}
\subsection{Lateral inhibition layer}

An intuitive way to explain the individual activations in hypercolumns of a module is to order them by their strength. The maximum activation value for a given input should correspond to the neuron that best matches a feature pattern according to the weights in the kernel. This behaviour is not ensured by the way the model parameters are updated, which is based on their backpropagated contributions to the output loss. Many neurons can have equal responses to the same input feature creating redundancy. A mechanism that ensures that a winner neuron amplifies its activation and inhibits others, is known in Neuroscience as \textit{lateral inhibition}. In a group of co-adapting neurons the response of one neuron antagonizes the response of others on the same stimuli. Our work facilitates this mechanism into the gradient descent learning process by adding a new Lateral Inhibition Layer (LIL) after the linear transformation that is performed by a convolutional layer. The lateral inhibition function is leveraging the properties of the softmax function \cite{gao_properties_2018} to implement the antagonism as:
\begin{equation}
    \label{eq:4}
    \textbf{z}^{(l)}_{x,y} = f_{LI}(\textbf{a}^{(l)}_{x,y}) = \textbf{a}^{(l)}_{x,y}(1 + \softmax{\textbf{a}^{(l)}_{x,y}})
\end{equation}

We use the LIL inside residual and dense modules after the last convolution operation and before any non-linear or normalization layer. Their respective transformation functions become:
\footnotesize
\[
\zres{(l)}{\zact{k-2}{X}}=\relub{\zli{\zconv{k}{ \relu{ \zconv{k-1}{\zact{k-2}{X}} } }} + \zact{k-2}{X} } 
\]\[
\zdens{(l)}{\zact{k-1}{X}}=   \zli{\zconv{k}{\relu{\zact{k-1}{X}}}}
\]
\normalsize

\setlength{\belowcaptionskip}{-5pt}
\begin{figure}[!h]
    \centerline{\includegraphics[width=4.5cm]{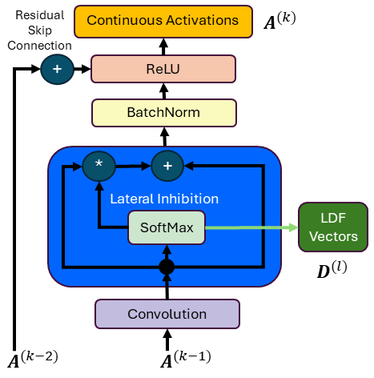}}
	\caption{Lateral Inhibition Layer (LIL) placement inside a residual module.}
	\label{fig:lil}
\end{figure}

\textbf{Lemma 1 - Lateral inhibition via amplification of gradients.}
For any pair of neurons in a hypercolumn, the lateral inhibition function (\ref{eq:4}) will amplify the gradients of the winner neuron, forcing larger updates in its weights in comparison to the others.

We present the proof of Lemma 1 in Section \ref{sec:appA} of the Appendix.

\paragraph{Conjecture 1.}
Increased weight regularization is needed for positively monotonic behaviour of the gradient amplification factors $\beta$ and $\gamma$ of the lateral inhibition function, and/or to restrict the input of the lateral inhibition function in the range $a_i \in [-6,6]$.

\subsection{Local discrete feature vectors} The output of a LIL is a continuous activation tensor $\textbf{Z}^{(l)} \in \imagespace{h}{w}{c_{out}}$ of spatial dimensions $h \times w$, where a patch descriptor slice $\textbf{z}^{(l)}_{x,y} \in \mathbb{R}^{c_{out}}$ has a corresponding local discrete feature (LDF) vector $\textbf{d}^{(l)}_{x,y}=\argsort{\softmax{\textbf{a}^{(l)}_{x,y}}} \in \mathbb{N}^{c_{LDF}}$ that is determined by the softmax output, with a length $c_{LDF} < c_{out}$ potentially less than the count of neurons in the hypercolumn. The learning process ensures that the scores of the softmax function are corresponding to the prevalence of a neuron over others, that is ordering their selectivity to an input feature. Thus, keeping the indices of the top ranked neurons can be used to explain the input space. Even though the number of ordered permutations could explode, experiments revealed that only a tiny fraction of these occur in a trained neural network predictor; this was expected since learning reduces uncertainty and removes randomness.

\subsection{Explanation process}
\begin{figure}[!h]
    \centerline{\includegraphics[width=14.2cm]{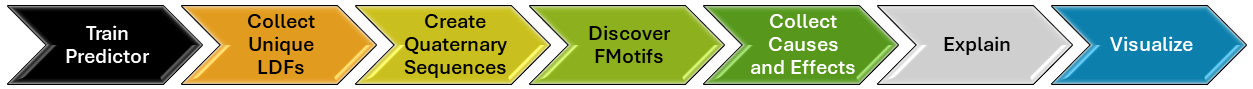}}
	\caption{Steps of the ExplaiNet framework process for the generation of explanations.}
	\label{fig:pipeline}
\end{figure}

\textbf{Supervised classification training:}
The process of generating explanations for an ExplaiNet classifier involves two training phases, each one followed by a collection phase.  The process starts with the supervised learning phase, that trains, or fine-tunes, a predictor on the available training samples as it is typically done. In this work we use standard minibatch Stochastic Gradient Descent (SGD) with momentum, while other optimizers can be used. During the first collection process, we recall all training set samples through the trained model, so that it generates LDF vectors for each image patch $x,y$ of each level $l$. This phase determines a set of unique LDF vectors $V^{(l)}=\{\zldfvector{l}{}(i)\}$, that constitutes a vocabulary of visual words \cite{sivic_video_2003}.

\textbf{Feature motif discovery:}
The second training phase employs unsupervised learning with EM. In the prepossessing step, we convert the vectors into the quaternary system so we can utilize any existing motif discovery algorithm. We have used YAMDA \cite{quang_yamda_2018}, that is an accelerated implementation of EXTREME \cite{quang_extreme_2014}. We run the algorithm on each set \footnotesize $V^{(l)}_{4}=\{\zldfvector{l}{4}(i)\}$ \normalsize of quaternary LDF sequences with a predetermined value for the basic hyperparameter \footnotesize $K_{motifs}$ \normalsize, that is the maximum count of motifs to discover. The process may stop early to a lower count, if there are no more significant patterns in the data. 
Before the start of the next iteration, the algorithm removes the training samples where the motif is present, by converting discrete $\zldfvector{l}{4}(i)$ into one-hot encoding matrices $\textbf{D}_{4}(i)$, and matches these with the motifs' PPMs. The matching function was changed from the original implementation  of YAMDA into the normalized 2D signal cross correlation $\rho({\cdot})$ of the rectified odds ratio between a motif's PPM $\zfmotifppm{l}{j}$ and the background probabilities $\textbf{B}$, of quaternary digits' occurrence at a position. The 2D discrete signal energy function is $\energy_{LDF}(i)= \sum^{c_{4}}_{n}{ \sum^{4}_{m}|d_4(i)[n,m]|^{2}}$ where the length of the quaternary vector is $c_{4}=c_{LDF}*log_{4}(c_{out}) \in \mathbb{N}$, for an LDF vector of length $c_{LDF}$ that is a multiple of 4, and for $c_{out}$ possible discrete of neuron indices.
\begin{equation}\label{eq:5}
\footnotesize
\nolinebreak{
f_{match}(\textbf{D}_{4}(i),\zfmotifppm{l}{j})= \frac{1}{\sqrt{\energy_{LDF}(i) \,\energy_{PPM}(j) }} \rho( \textbf{D}_{4}(i),  \relu{\log{\zfmotifppm{l}{j}} - \log{\textbf{B}}} )
}
\normalsize
\end{equation}

The feature motif (FMotif) discovery phase drastically reduces the size of the LDF vocabulary into a new vocabulary $M^{(l)}$ of size \footnotesize $N_{motifs} \leq K_{motifs} \leq |M^{(l)}| \ll |V^{(l)}|$. \normalsize This serves conciseness \cite{parekh_framework_2021} that is one of the required characteristics for comprehensible explanations. Low values of \footnotesize $K_{motifs}$ \normalsize can be used for extremely concise explanations, using just a handful of FMotifs. We load the vocabularies $M^{(l)}$ to corresponding levels of our explainer and it can infer a scalar FMotif index from a non-explainable hypercolumn activation vector. Thus, the explainer provides a discrete image representation matrix $\boldsymbol\Psi^{(l)} \in \matrixspacediscrete{h}{w}$ for each intermediate level of the network.

\textbf{FMotif effects and causes collection:} 
 \cite{parekh_framework_2021}. 
In the second collection phase we recall through ExplaiNet samples that were not part of the training set, to make observations. The LDFs are matched with FMotifs using the matching function (\ref{eq:5}). For each node $o^{(l)}_{x,y}$ of the explainer graph, we record the FMotif index $m^{(l)}(o^{(l)}_{x,y})$ as the effect observed on image patch that is presumably caused by the set of effects in the backward adjacent nodes. For CNN networks, the causes $C^{(l-1)}_{x,y}=\{m^{(l-1)}(o^{(l-1)}_{x-i,y-j})\}: {i,j \in [-\frac{n}{2}, +\frac{n}{2}]\}}$ correspond to the ${n}\times{n}$ receptive field of the convolution operation. This is our ground truth observation that we keep in pairs $\{m^{(l)}_{x,y}, C^{(l-1)}_{x,y}\}$ for evaluating the quality of explanations. There are edges to "null" nodes in the graph that stand for zero padding, and are not included in the set of causes since batch normalization layers perform zero-mean centering. Also, pairs of $\{\boldsymbol\Psi^{(l)}(i), \hat{y}_{i}\}$ are kept for evaluating explanations on the model predictions, with $\hat{y}_{i}$ the predictor's class index for sample $i$. The collection process creates Pareto histograms for cause sets and image representations, where bins correspond to FMotifs ordered by their frequencies. Their ordered indices are kept in vectors $\textbf{c}^{(l)}_{x,y}$ and  $\pmb{\boldsymbol\psi}^{(l)}(i)$.

\textbf{Bayesian network explanations:} 
The directed acyclic explainer graph expresses the behaviour of the convolutional moving window that draws edges between patch nodes. These are considered causal links, since the activation value of a trained neuron is influenced by the values in its receptive field in a deterministic manner \cite{lacave_review_2002}. Hence we consider our explainer a Bayesian network, a model with inherent explainability. We define as $X$ the event of an FMotif presence in one or more patches of the image representation $\boldsymbol\Psi^{(l)}$, or equally having a non-zero count in the histogram. During the collection phase we count occurrences of FMotif $m_i$ and calculate its marginal probability 
$P(m_i)=\frac{ \zcount{X=m_i} }{q}$ where $q$ the total count of occurrences. For an outcome event $Y$ "... therefore the prediction is explained as $\explained{y}_c$" the prior probability is $\explained{y}_c=\frac{ \zcount{Y=\explained{y}_c} }{q}$. The conditional probability for an FMotif presence $X$ to be the cause of outcome $Y$ is $P(m_i | \explained{y}_c)=\frac{ \zcount{X=m_i\,Y=\explained{y}_c} }{q}$. It is easy to think that the presence of one FMotif index in the histogram depends on the convolution operation at a specific location. Since it moves across the image with the same kernel the presence of a second FMotif index in the histogram is conditionally independent. Thus, we can use a Naive Bayes classifier to make a surrogate prediction based on FMotif indices, that is our stochastic explanation. This provides us with the three basic aspects of an explanation; i) why class $c$ was predicted, ii) why class $c^{'}$ was not predicted instead, and iii) what to expect for an unknown sample, based on the explained behaviour of the model on some samples. The explained class index at an intermediate level $l$ is:
\begin{equation}\label{eq:6}
\footnotesize
\explained{y}_c = \argmax{c}{P(\explained{y}_c)\prod^{N^{(l)}_{motifs}}_{i} P(m_i | \explained{y}_c) }
\normalsize
\end{equation}

\label{others}
\section{Experiments and results}
\subsection{Prediction}
\textbf{Datasets and augmentation:} We train our models from scratch on the MNIST dataset \cite{lecun_gradient-based_1998}, Fashion MNIST (FMNIST) that contains tiny grayscale images of 10 fashion items \cite{xiao_fashion-mnist_2017}, Kuzushiji MNIST (KMNIST) that has Japanese cursive syllabograms with 10 classes, one for each row of Hiragana syllabary \cite{clanuwat_deep_2018}, Oracle MNIST (OMNIST) that depicts ancient Chinese ideograms of 10 concepts and the tiny color image benchmark dataset CIFAR10 \cite{krizhevsky_learning_2009}. Our approach for the training data feed that ensures deterministic training and fair comparison is described at Section \ref{sec:appB} of the Appendix.

\textbf{Supervised training experiments:} We use LIL in ResNet and DenseNet to derive R-ExplaiNet and D-ExplaiNet. For each ResNet pair we train 10 random folds, and for DenseNet pairs 5 due to their increased complexity, for at least 8 different feature-layer combinations per dataset. Each model was trained on a single GPU, where details on hyperparameters, our software implementation that will become available on GitHub, and infrastructure can be found at Appendix \ref{sec:appC}. For fair comparison with the baseline, data and training hyperparameters are kept the same for all models of a task, except $\lambda$ of weight decay regularization. Our formal proof suggests that ExplaiNets need higher values for gradient stability; that was confirmed with our first experiments on MNIST, where we tried at least 9 models with combinations of constant features and layers, and a reverse pyramid of features.

\setlength{\tabcolsep}{2pt} 
\begin{table}[!h]
    \centering
    \caption{Average error rate over 10 folds within 95\% confidence interval by feature setup on MNIST, using $\lambda=0.001$ for all models. R-ExplaiNet-C restricts input values of LILs to [$-6,6$].}
    
\begin{tabular}{ccccccccc}
        \toprule
\footnotesize{Id} & \multicolumn{2}{c}{R8} & \multicolumn{2}{c}{R16} & \multicolumn{2}{c}{R32} & \multicolumn{2}{c}{R64} \\
        \cmidrule(r){2-3}
        \cmidrule(r){4-5}
        \cmidrule(r){6-7}
        \cmidrule(r){8-9}
        \footnotesize{K} & ResNet & R-ExNet-C & ResNet & R-ExNet-C & ResNet & R-ExNet-C & ResNet & R-ExNet-C \\              
        \midrule
18 & 0.562\ci{0.03} & \underline{0.546}\ci{0.04} & 
0.387\ci{0.02} & \underline{0.364}\ci{0.03} & 
0.323\ci{0.02} & \underline{0.311}\ci{0.01} & 
\underline{0.287}\ci{0.01} & \underline{0.287}\ci{0.02} \\
22 & 
\underline{0.462}\ci{0.02} & 0.513\ci{0.03} & 
\underline{0.344}\ci{0.02} & 0.346\ci{0.02} & 
\underline{0.275}\ci{0.02} & 0.283\ci{0.03} & 
0.263\ci{0.01} & \textbf{\underline{0.256}}\ci{0.01} \\
26 & 0.488\ci{0.03} & \underline{0.469}\ci{0.03} & 
\underline{0.328}\ci{0.03} & 
0.354\ci{0.03} & \underline{0.302}\ci{0.02} & 
0.294\ci{0.02} & \underline{0.273}\ci{0.02} & 
0.292\ci{0.02} \\
        \bottomrule
      \end{tabular}
      
\end{table}

The MNIST classification task could be considered as solved; nevertheless it still serves the purpose of providing a handful of misclassified samples that can be investigated via explanations. The difference in performance lies in the second decimal digit of error rate percentage, that is 1-10 validation samples. The minimum error amongst all MNIST models was $0.200$ for R-ExplaiNet22-64 of \modelsize{0.75} parameters using $\lambda=0.001$, and its average error stabilizes to 0.258\ci{0.008} for $\lambda=0.002$. Additionally, we experimented with a second type of LIL that clips its input values to be inside [$-6,6$] that resulted in the overall lowest average error of 0.256\ci{0.013}.
In FMNIST, KMNIST, OMNIST and CIFAR10, where overfiting is exacerbated, R-ExplaiNets consistently surpass the baseline accuracy in every setup. Also, experiments reveal that accuracy of DenseNets increases when LIL is used in their modules. A complete set of experiments is included in Appendix \ref{sec:appD}.

\setlength{\tabcolsep}{2pt} 
\begin{table}[!t]
    \centering
    \caption{Average accuracy for top performing residual networks in FMNIST, KMNIST, OMNIST, CIFAR10 with parameter size 0.89MP.}
    \begin{tabular}{lcccccc}
        \toprule    
          Model
          & Fashion MNIST
          & Kuzushiji MNIST
          & Oracle MNIST 
          & CIFAR10 
          & MNIST \\          
        \midrule    
          R-ExplaiNet26-64
          & \textbf{93.03}\ci{0.119}
          & \textbf{98.66}\ci{0.049}
          & \textbf{96.68}\ci{0.109}
          & \textbf{93.80}\ci{0.119} 
          & 99.70\ci{0.022}
          \\
          ResNet26-64
          & 92.83\ci{0.131}
          & 98.53\ci{0.061}
          & 96.56\ci{0.123}
          & 93.41\ci{0.093}
          & \textbf{99.73}\ci{0.015}
          \\
        \bottomrule         
    \end{tabular}
\end{table}

\setlength{\tabcolsep}{2pt} 
\begin{table}[!t]
    \centering
    \caption{Accuracy of CIFAR10 classifiers that were converted into ExplaiNets.}
    \begin{tabular}{lcccccc}
        \toprule    
           ~
          & \footnotesize{Res20 (R1)}
          & \footnotesize{R-ExN20 (R1)}
          & \footnotesize{Dens40 (R12)}
          & \footnotesize{D-ExN40 (R12)}
          & \footnotesize{Dens100BC (R12)}
          & \footnotesize{D-ExN100BC (R12)} \\
        \midrule    
          Acc.
          & 91.59\ci{0.117}
          & \underline{91.88}\ci{0.133}
          & 93.41\ci{0.152}
          & \underline{93.8}\ci{0.08}
          & 94.38\ci{0.182}
          & \textbf{\underline{94.6}}\ci{0.163}
          \\       
        \bottomrule         
    \end{tabular}
\end{table}

\textbf{Efficiency assessment:} We use our RME index for average accuracy metrix, to report the most efficient predictor for each task, its accuracy and size. We include in the group only models above certain accuracy threshold. ExplaiNets can achieve a sufficient level of performance for a task with a few parameters. More details on each individual model efficiency are included in Appendix \ref{sec:appD} 

\setlength{\tabcolsep}{4pt} 
\begin{table}[!ht]
    \centering
    \caption{Most efficient models above an accepted performance limit for each task. }     
    \begin{tabular}{lcccccc}
        \toprule    
          ~
          & Fashion MNIST
          & Kuzushiji MNIST
          & Oracle MNIST 
          & CIFAR10 
          & MNIST \\
          
        \midrule    
        Acc. Limit
        & 92.0
        & 97.0
        & 95.0          
        & 92.0                 
        & 99.5 \\  
         Model   
        & R-ExN22-48
        & R-ExN18-16
        & R-ExN18-32         
        & R-ExN18-48             
        & R-ExN22-16 \\
        RMEF
        & 1.30
        & 2.19
        & 1.74         
        & 1.27             
        & 2.58 \\
        Accuracy
        & 92.86\ci{0.13}
        & 97.63\ci{0.08}
        & 96.13\ci{0.13}         
        & 92.61\ci{0.10}             
        & 99.67\ci{0.03} \\
        Size (MP)
        & 0.420
        & 0.038
        & 0.150         
        & 0.337             
        & 0.048 \\

        \bottomrule         
    \end{tabular}
\end{table}
\begin{figure}[!h]
    \centering
    \includegraphics[width=0.30\textwidth]{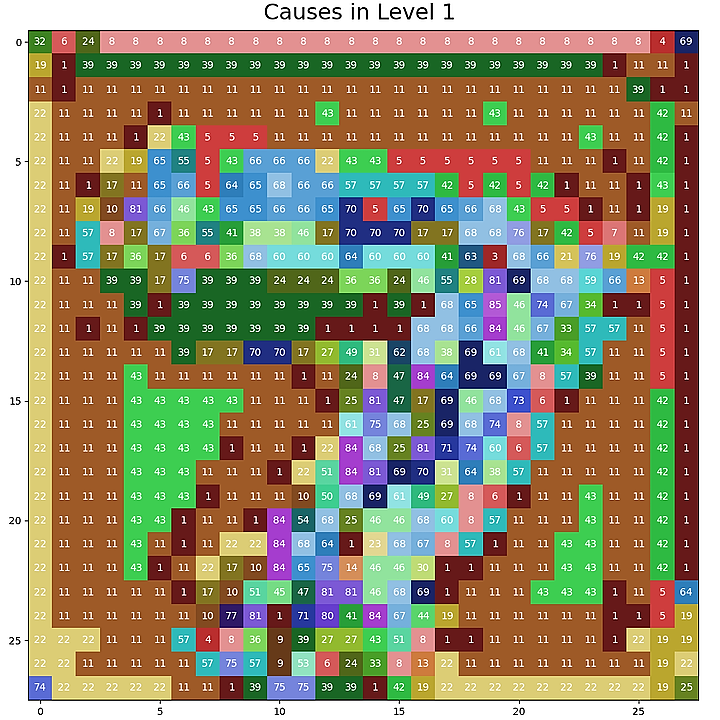}
    \includegraphics[width=0.26\textwidth]{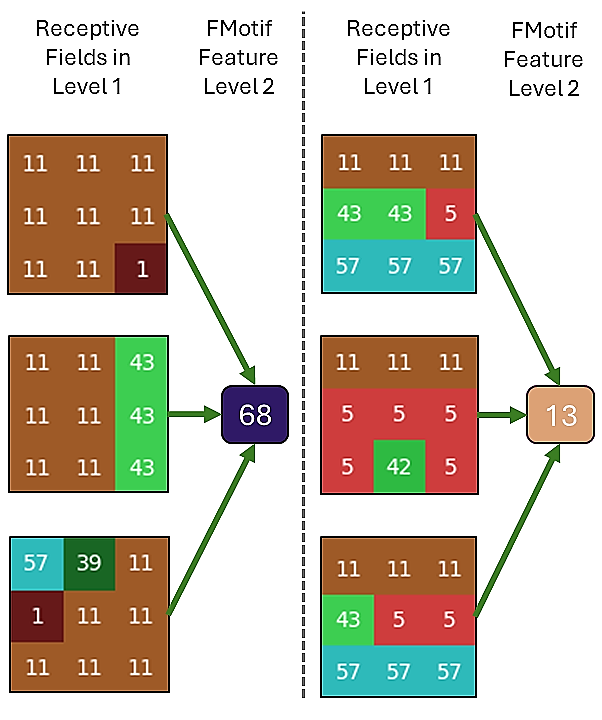}
    \includegraphics[width=0.30\textwidth]{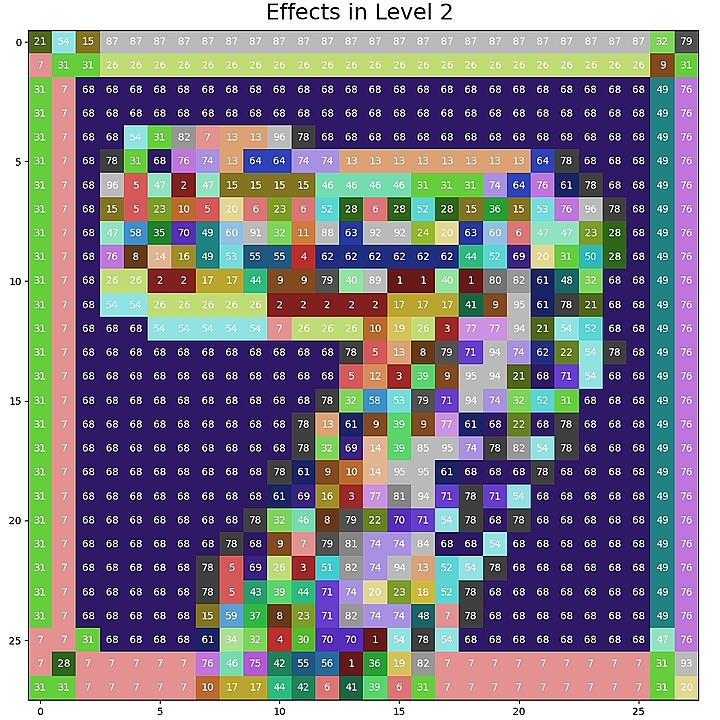}
    \caption{Discrete feature mosaics. Left:\nth{1} level mapped to a \footnotesize{$7\times7$}\normalsize\,area of the input image, right:\nth{2} level to \footnotesize{$11\times11$}\normalsize. Middle: Observed causes $C^{(1)}_{x,y}$ in the receptive field of effects $m^{(2)}_{68}$ and $m^{(2)}_{13}$.}
    \label{fig:motif_frescos}
\end{figure}

\subsection{Interpretations and Explanations}
\textbf{Intepretations:} We run the explanation process for R-ExplaiNet18-16, a residual ExplaiNet of 18 layers and 16 features per layer trained on MNIST, with $K_{motifs}=96$, to discover FMotifs for 8 levels of explanations on the unique LDFs in the training set. The discrete representation matrices are visualized as mosaics, depicted in figure \ref{fig:motif_frescos}. Matching scores of FMotifs can be used to generate heatmaps, that illustrate how they are selective to different classes, as presented in figure \ref{fig:heatmaps}. Full hyperparameters of discovery and extra analysis on interpretations are available in Appendix (\ref{sec:appE}).

\setlength{\belowcaptionskip}{-15pt}
\begin{figure}[!h]
    \centering
    \includegraphics[width=0.5\textwidth]{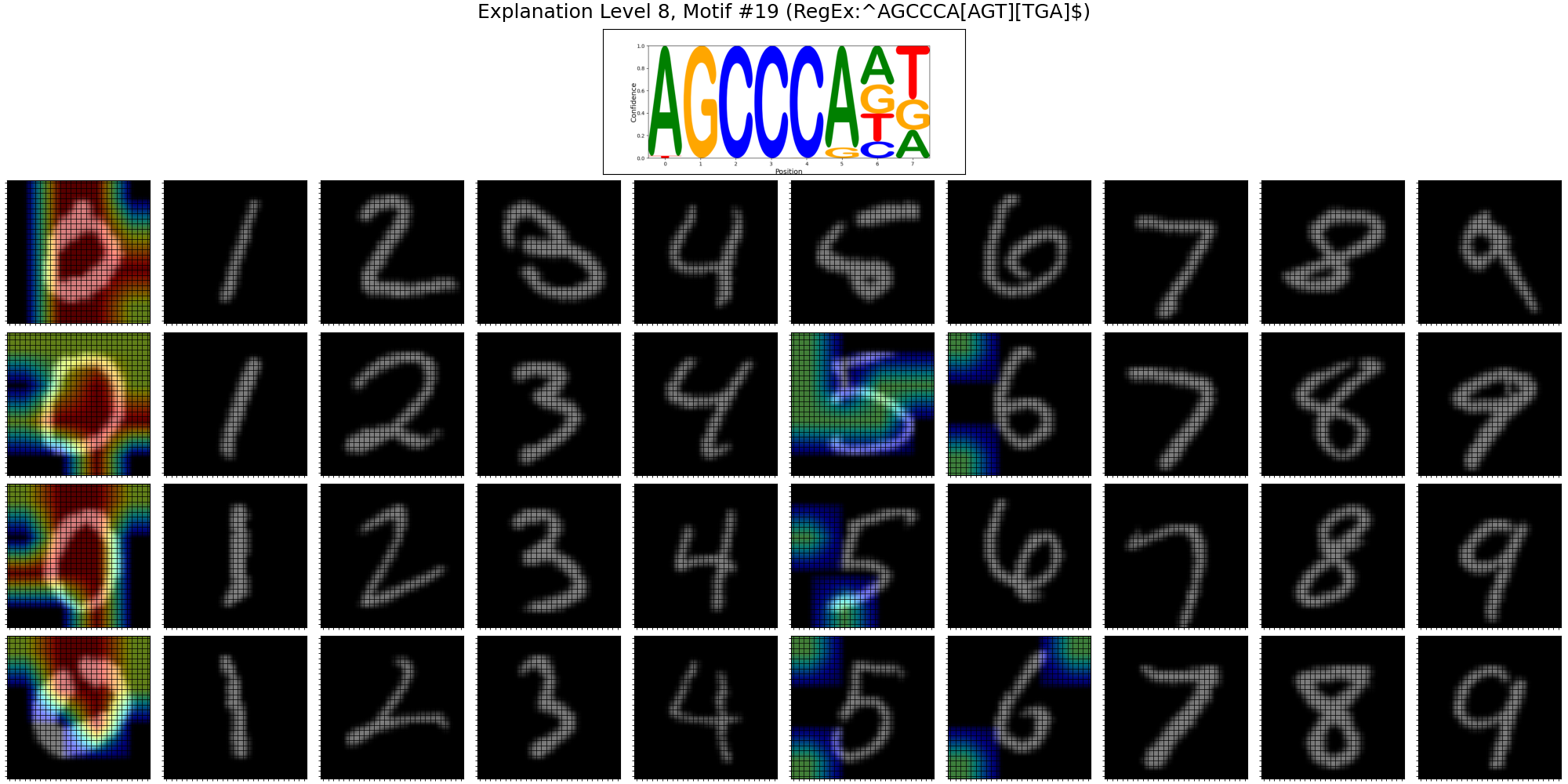}
    \caption{Matching scores superimposed on the input image for FMotif 19 of level 8 in R-ExplaiNet18-16, along with its logo. At this explanation level we have global feature motifs in a \footnotesize{$4\times4$} matrix.}\normalsize
    \label{fig:heatmaps}
\end{figure}


\textbf{Explanations:}
We recall all samples of the MNIST validation set through R-ExplaiNet18-16, to collect Pareto histograms of FMotif representations and image slices, calculating probabilities. For each level of explanation, the explained class is inferred based on the presence of FMotifs in representations, and FTO is calculated. The results show that local FMotif features are not faithful to the output of the model, while global are. The FCE is calculated for all FMotif in the vocabulary and for the top 20\% of them. Details about them can be found at Appendix \ref{sec:appF}.
The CNN predictor's accuracy is $99.58$, that can be compared to surrogate predictions from all levels.

\setlength{\tabcolsep}{2pt} 
\begin{table}[!ht]
    \centering
    \caption{Explainability metrics for R-ExplainNet18-16 on the MNIST validation set.}
    \begin{tabular}{lccccccccc}
        \toprule    
            ~
          & Metric
          & Level1 & Level2 & Level3 & Level4 & Level5 & Level6 & Level7 & Level8 \\
        \midrule    
          FMotif Vocab.
          & Count
          & 96
          & 85
          & 96
          & 85
          & 96
          & 73
          & \textbf{58}
          & 60\\         
          Surrogate Pred.
          & Acc.
          & 9.92
          & 10.02 
          & 17.42
          & 18.00
          & 53.09
          & 81.04 
          & 98.21 
          & \textbf{99.50}\\
          Explain Pred.
          & FTO
          & 9.80
          & 9.92
          & 17.31
          & 17.86
          & 52.99
          & 80.94
          & 98.23
          & \textbf{99.62}\\       
          All effects          
          & FCE
          & ~ 
          & \textbf{92.5}\ci{1.6}
          & 70.0\ci{3.2}
          & 92.4\ci{0.9}
          & 88.9\ci{1.4} 
          & 98.7\ci{0.3} 
          & 88.5\ci{2.0} 
          & 69.3\ci{3.8}\\     
          Most explained
          & FCE\
          & ~ 
          & 99.6\ci{0.1}
          & 88.7\ci{1.3}
          & 97.5\ci{0.5}
          & 95.7\ci{0.4}
          & \textbf{99.7}\ci{0.1} 
          & 96.9\ci{0.9} 
          & 88.1\ci{3.5}\\            
          Best FMotif
          & FCE
          & ~ 
          & 99.96 
          & 94.93 
          & 99.76
          & 97.67 
          & \textbf{100.00}
          & 99.70 
          & 99.46\\ 
          
        \bottomrule         
    \end{tabular}
\end{table}


\section{Discussion}
\subsection{Key findings}
Experiments clearly indicate that our explainable-by-design CNN predictor retains or surpasses performance in comparison to the baseline, while it is able to provide explanations based on discrete features. Training the best performing R-ExplaiNet on MNIST with increased weight decay revealed that doubling the amount of regularization increases the mean accuracy.The reproducible error rate of the best 
model is $0.2$\% which is comparable to the current state-of-the-art for single models. This shows promise for reaching the highest performance on tasks, if more complex models and elaborate training schemes are used.

The feature motif discovery process underutilizes the learning capacity of EXTREME, because our genome-encoded LDFs are aligned sequences, while the algorithm can find motifs in the middle of sequences. Keeping the conversion to quaternary numeral system ensures a logarithmic space complexity $\mathcal{O}(m \cdot log_{4}(n))$ for features $n$ and length $m$ of LDFs, thus the method can scale up to an extremely high count of features for short LDFs, or have a length equal to the count of features. 

Our explainer graph, that infers an FMotif as the integer scalar image patch descriptor, can generate online explanations. We have performed a detailed investigation of samples that are unfaithful to the predictors output, and those that are misclassified by both the CNN and the Bayesian explainer. We have noticed interesting cases where the neural predictor fails but the explainer is correct. This creates the potential of using explanations from the last levels together with the neural predictor in an ensemble, or using the explainer as a detector of adversarial samples.

Evaluation of fidelity to output for the first levels indicates that, even though the intermediate non-explainable representations arithmetically lead to the output prediction, the presence of a specific feature is not a causal reason for it. These start to emerge as the network growths in depth and the receptive field of neurons expands. When explaining intermediate feature interactions in unknown samples, there are 20\% FMotif effects that are sufficiently explained with high FCE, thus their presence at a level is consistently faithful to the observed FMotif causes of the previous level.

\subsection{Limitations and future work}
Further research on regularization aspects of the LI function is required, to investigate the magnitude of weights and activations during training. The proper amount of weight decay and restriction of input values for positively monotonic gradient amplification needs to be further investigated. Numeric stability seems to play an role here and training models with fp64 numbers will provide insight. Variety and amount were preferred over scale for our limited computational resources, thus we have not trained our models on medium resolution images, high number of classes, nor used models with millions of parameters. The next version of the software implementation will parallelize the collection phases, enabling multiple comparative explanation experiments. The EXTREME algorithm could be adjusted to support custom numeric system bases, so that a conversion to the quaternary system will become an optional decision, to be taken as a mitigation of high complexity.

\section{Related Work}
Our work falls under the general scientific field of XAI\cite{gilpin_explaining_2019} that has a large taxonomy of sub-fields where ExML and/or statistical learning is used to provide interpretability and explainability. For neural networks we can identify the main branches of: Surrogate (proxy) Models, Automatic Rule Extraction, Attribution Methods, Example-Based Methods and Explainable-by-Design Neural Networks. Neural network surrogate explanations are based on model-agnostic methods that explain a black-box model \cite{ribeiro_model-agnostic_2016} or use a decision tree as a surrogate \cite{pedapati_learning_2020} \cite{liu_learning_2021}. Rules can be extracted with various approaches including those that follow programming logic\cite{yang_learning_2021}\cite{trivedi_learning_2021}. Attribution methods, that mostly provide interpretability, is a diverse branch. Beyond gradient-based attribution methods \cite{selvaraju_grad-cam_2017}\cite{shrikumar_learning_2017}\cite{smilkov_smoothgrad_2017}\cite{li_deep-lift_2022} and attribution based on Shapley-values \cite{lundberg_unified_2017}\cite{chen_algorithms_2023} there are axiomatic attributions\cite{sundararajan_axiomatic_2017}\cite{hesse_fast_2021}, perturbation-based methods \cite{ribeiro_model-agnostic_2016}\cite{ivanovs_perturbation-based_2021}, rule-based attributions \cite{letham_interpretable_2015}, information-bottleneck attributions \cite{zhang_fine-grained_nodate} and attention saliency maps \cite{shitole_one_2021}. The requirements of a proper explanation \cite{faye_explanation_1999} are better satisfied by the Example-Based Methods, where explanations can be based on counterfactuals \cite{laugel_achieving_2023}\cite{hamman_robust_2023} or influential instances \cite{koh_understanding_2017}. Many proposed methods that provide explainability through the use of prototypes \cite{hampton_concepts_2006} can be found in the literature \cite{bien_prototype_2011}\cite{alvarez_melis_towards_2018}\cite{li_deep_2018}\cite{papernot_deep_2018}\cite{hase_interpretable_2019}\cite{crabbe_explaining_2021}\cite{kim_xprotonet_2021}\cite{nauta_neural_2021}.

More related to our work are explainable-by-design neural networks, like additive models \cite{agarwal_neural_2021}, dynamic alignment networks \cite{bohle_convolutional_2021} and neural decision trees \cite{olaru_complete_2003}\cite{irsoy_soft_2012}\cite{frosst_distilling_2017}\cite{ojha_backpropagation_2022}\cite{zhao_preliminary_2022}. The ExplaiNet is mostly suited under the branch of Hybrid Explainable-by-Design Neural Networks like the work in \cite{alvarez_melis_towards_2018} from which we embrace the concept of predictor-explainer working in pair. Nevertheless, LDFs and FMotifs provided by our framework can be used by other branches of ExML.

Considering the use of softmax as an activation function, we utilize its characteristics for a different purpose when compared to the Attention layer \cite{lin_survey_2022}\cite{khan_transformers_2022}, that uses it as an activation function over the inner product of query and key, scores of which are used to match a value. When used in our LIL, softmax steers the gradient descent learning rule toward incorporating lateral inhibition. It can be considered as closely related with the long-standing concept of Soft Competitive Learning \cite{nowlan_maximum_1989}, but our model does not use a learning rule that belongs to Hebbian learning, when compared to the closely related work of \cite{nicosia_training_2022}.


\section{Conclusions}
In this work, we present a novel framework for explainability and interpretability that considers using discrete features to serve their needs. An explainable-by-design CNN learns local discrete feature vectors that indicate a ranked matching of neuron-specific weights to the input receptive field. These belong to the convolution kernel that is learned with supervised gradient descent. We can replace discrete vectors with a single scalar discrete value of a feature motif, for image patches at all intermediate image representations in the network. The proposed Lateral Inhibition Layer is generally applicable in neural networks, and we have experimentally proven that our novel ExplaiNet predictor retains the accuracy of its baseline architecture, breaking the performance-explainability trade-off. Our explainer graph provides Bayesian explainability for understanding causes in an explanation level that lead to the occurrence of a feature in the next level, and can be also used for surrogate class prediction from all levels. Our framework comes with a complete proof-of-concept software implementation that can be used in future research efforts towards explainability.

\begin{ack}
\section{Author Contributions}

The primary author conceived the idea for the method and the overall framework, based on earlier work done on CNN-based feature quantization, for the task of content-based image retrieval. The second author supervised the project, setting the experimental requirements, advising, and providing valuable feedback. The code for the software implementation was written by the primary author and the manuscript was written by the first author and edited by the second.


This work utilized compute resources at the Department of Information and Electronic Engineering, International Hellenic University (IHU), at Sindos. The authors would like to thank Professor Antonis Sidiropoulos for his valuable work for procuring and operating this equipment and his support during the experiments phase.

Pantelis I. Kaplanoglou received funding by the MANOLO EU grant at the time of this work completion.
\end{ack}


\bibliographystyle{IEEEtran}
\medskip

{
\small
\bibliography{aReferences}
}

\appendix

\newpage
\section{Appendix / Proof of lateral inhibition via amplification of gradients}
\label{sec:appA}

\paragraph{Lemma 1 - Lateral inhibition via amplification of gradients.}
For any pair of neurons in a hypercolumn, the lateral inhibition function 
\begin{equation}
    \textbf{z}^{(l)}_{x,y} = f_{LI}(\textbf{a}^{(l)}_{x,y}) = \textbf{a}^{(l)}_{x,y}(1 + \softmax{\textbf{a}^{(l)}_{x,y}})
\end{equation}

will amplify the gradients of the winner neuron, forcing larger updates in its weights in comparison to the others.

\paragraph{Proof:}
Let $\textbf{x}$ the input vector, that is the flattened receptive field matrix, $\textbf{a}$ the hypercolumn's activation vector, $i$, $j$ indices of two neurons and $\textbf{W}$ the shared weight kernel. The 2D convolution is the function $\textbf{a} = g(\textbf{x};\textbf{W})$ and its partial derivatives w.r.t to the weights are $\nabla g(a_i)=\frac{\partial}{\partial \textbf{W}}{a_i}$. 

The lateral inhibition function centered at position $x,y$ for a neuron $i$ is $f(i,\textbf{a}) = a_i + a_i  s(a_i, \textbf{a})$, where $s(a_i, \textbf{a})$ is the softmax function with neuron $i$ in the numerator and other neurons in the denominator.

\begin{align}
\nabla f(i, \textbf{a}) 
= \frac{\partial}{\partial \textbf{W}}{a_i} + \frac{\partial}{\partial \textbf{W}}{a_i s(a_i, \textbf{a})} \\
=\frac{\partial}{\partial \textbf{W}}{a_i} + s({a_i}, \textbf{a}) \frac{\partial}{\partial \textbf{W}}{a_i} + {a_i} \frac{\partial}{\partial\textbf{W}}{s(a_i,\textbf{a})} \\
=\frac{\partial}{\partial \textbf{w}}{a_i} + s({a_i}, \textbf{a}) \frac{\partial}{\partial \textbf{w}}{a_i} + {a_i}\frac{\partial s(a_i,\textbf{a})}{\partial a_i}  \frac{\partial {a_i}}{\partial \textbf{w}} \\ =\zpartderivw{a_i \biggl(1 + s(a_i,\textbf{a}) + a_i\zpartderiva{ s(a_i,\textbf{a})} \biggr)}
\label{eq:eq_PartialDerivatives}
\end{align}

The softmax function for two input neurons $i$,$j$ of the same hypercolumn is $\softmax{a_i}=\softmax{a_{i},\textbf{a}}$ where neuron $i$ is in the numerator. The element $i$,$j$ of the Jacobi matrix for the softmax function is:

\begin{align}
\label{eq:eq_JacobiSoftmax}
\nolinebreak{J(i, j, \textbf{a})=\zpartderiva{ s(a_i, \textbf{a})}=\zpartderiva{ s(a_i)}=s(a_i)( \iverson{i=j} - s(a_j))}
\end{align}

where $\iverson{.}$ is the Iverson bracket. We combine (\ref{eq:eq_PartialDerivatives}) and (\ref{eq:eq_JacobiSoftmax}) into

\begin{align}
\nabla f(i, j, \textbf{a})=\zpartderivw{a_i \biggl(1 + s(a_i) + a_is(a_i)( \iverson{i=j} - s(a_j)) \biggr)}\\
=\zpartderivw{a_i \biggl(1 + s(a_i)(1 + a_i( \iverson{i=j} - s(a_j)) \biggr)}
\label{eq:eq_CombinedDerivatives}
\end{align}

We plug the derivative of the 2D convolution into ((\ref{eq:eq_CombinedDerivatives})) for the case ${i}\neq{j}$ it becomes
\begin{align}
\nolinebreak{\nabla f(i, j, \textbf{a}) = \nabla g(a_i)\biggl(1 +  s(a_i) - {a_i}s(a_i)s(a_j)\biggr) =\beta \nabla g(a_i)}
\end{align}  
,while for the case $i=j$, of a diagonal element of the Jacobi matrix where $s(a_i)=s(a_j)$, it is 

\begin{align}
\nabla f(i, i, \textbf{a}) 
= \nabla g(a_i)\biggl(1 + s(a_i) (1 + a_i - a_is(a_i)) \biggr) \\
= \nabla g(a_i)\biggr(1 +  s(a_i) + a_is(a_i)-{a_i}s(a_i)^{2}\biggl) =\gamma \nabla g(a_i)
\end{align}

Considering that $s(a_i)=1-s(a_j)$ the limits below are zero
\begin{equation}
\lim_{s(a_i) \to 1}s(a_i)s(a_j) = \lim_{s(a_j) \to 1}s(a_i)s(a_j)= \lim_{s(a_j) \to 1}s(a_i)(1-s(a_j)) =0
\end{equation}
\newpage
For 
\begin{align}
\beta=(1 +  s(a_i) - {a_i}s(a_i)s(a_j))    
\end{align}
and
\begin{align}
\gamma=(1 +  s(a_i) + {a_i}s(a_i)-{a_i}s(a_i)^{2})
\end{align}  it is evident that

\begin{align}
\lim_{s(a_i) \to 1}\beta = 2 , \lim_{s(a_j) \to 1}\beta=1, 
\lim_{s(a_i) \to 1}\gamma = 2 , \lim_{s(a_i) \to 0}\gamma=1 
\end{align} which proves the amplification of the gradient for the candidate winner neuron $i$ in all elements of its Jacobi matrix.

\paragraph{Corolary 1.}
For $s(a_i) = s(a_j) =0.5$ we have the same adjustment of gradients, either amplification or attenuation.

We calculate:
\begin{align}
    \label{eq:eq_BetaForEqualActivation}
    \beta=1 + 0.5 - 0.25a_i=1.5 - 0.25{a_i}
\end{align}
\begin{align}
    \label{eq:eq_GammaForEqualActivation}
    \gamma=1 + 0.5 + 0.5a_i - 0.25a_i =1.5 + 0.25{a_i}
\end{align}
The adjustment of gradients, that is either amplification or attenuation, is the same for both neurons. 

Studying $a_i \in [-6,6]$, when $a_i=-6$ gradients of the non-diagonal elements in the Jacobi matrix will be amplified $\times 3$, while gradients in the diagonal will be inhibited to zero. The inverse is done for $a_i=6$, where non-diagonal gradients are zeroed, while gradients of the diagonal are amplified $\times 3$. 

\paragraph{Conjecture 1.}
Increased weight regularization is needed for positively monotonic behaviour of the gradient amplification factors $\beta$ and $\gamma$ of the lateral inhibition function, and/or to restrict the input of the lateral inhibition function in the range $a_i \in [-6,6]$.

When $a_i < -6$ or $a_i > 6$ we notice inversion of the gradient sign 
by either $\beta$ or $\gamma$. The unbounded nature of linear relations (\ref{eq:eq_BetaForEqualActivation}) (\ref{eq:eq_GammaForEqualActivation}) can lead to the exploding gradients problem, that we suspect it is mitigated by weight decay regularization combined with the use of batch normalization layers.


\section{Appendix / Training setup}
\label{sec:appB}

\subsection{Training data feed}

The training data feed is a pipeline of methods on the original samples of the dataset. They are pushed in a scheme that implements the iterator software design pattern; The model pops the next mini-batch of samples from the data feed iterator. In all our experiments the pipeline is:
\[
foreach(\textbf{S} \in TS) \to \text{std}(\textbf{S}) \to \text{aug}(\textbf{S} \to MB) \to \text{mix}(MB) \to \infty(MB \to \{\textbf{X}\}).
\] where $TS$ the training set, $std$ per channel image standardization, $aug$ reproducible random augmentation, $mix$ reproducible random shuffling on mini-batches of samples, $\infty(MB)$ an iterator on the mini-batches.

\subsection{Data preprocessing and augmentation}
We use a simple data augmentation and standardization scheme which follows standard approaches. We standardize with the per-channel mean $\mu_{TS}(c)$ and standard deviation {$\sigma_{TS}(c)$} for the samples of the training set,  where $c=1$ for grayscale images and $c \in \{r,b,g\}$ for color images.  

For MNIST, animage sample \textbf{X} is zero-padded with 3 pixels and a random $28 \times 28$ crop is taken. For FMNIST, the image is randomly left/right mirrored (horizontal flip) then it is padded with 3 pixels and a random crop is taken. For KMNIST and OMNIST,  we prefer the same scheme with MNIST since we consider left/right mirroring a non-label-preserving transformation of their image for writing symbols, e.g. numbers,letters,syllabograms,ideograms. For experiments on MNIST, FMNIST, KMNIST, CIFAR10 the size of the mini-batch is 128 samples while for OMNIST is 64 samples. The same random seed $rnd_{i}$ per fold index $i=1,...,10$ is used across all experiments.

\paragraph{Random number generators:}
In a Python implementation of a learning process there are several random number generators that need to be seeded to ensure determinism. We seed the value $rnd_{i}$ to:
\begin{itemize}
\item Generator of Python's \verb|random| package.
\item Python's hashing random generator.
\item \verb|numpy| package random number generator/
\item Tensorflow random number generator.
\item Keras random number generator.
    
\end{itemize}

All these provisions ensure deterministic reproduction of training by using the same sequence of input samples, and fair comparison between models by using the same initial random weights.

\section{Appendix / CNN architectures, software implementation and training infrastructure}
\label{sec:appC}

\textbf{Explainable-by-design CNN architectures:}
We have chosen two popular architecture of CNN for our experiments that they both implement skip connections to help the gradient flow towards the first layer. They are both counteracting the vanishing gradient problem for networks of increased depths. Another reason for the selection is the trade-off of having less parameters in a DenseNet, but more computations in comparison to a ResNet, a trade-off between space and time complexity of models. 

Placing LIL layers inside a ResNet of layer depth $K$ the levels of explanation for the R-ExplaiNet are $L=(K-2)/2$. 

When DenseNet is used as baseline, there are $B$ blocks each one containing $C$ modules. For the D-ExplaiNet $C$ LIL outputs are concatenated per block, while in an additional $B-1$ transition modules there is an LIL after the convolution and before the spatial downsampling operation. The count of available explanation levels for $K$ layers, is $L=K-2$. In the D-ExplaiNet the explainer graph nodes have edges to nodes that belong to previous levels $l-1, l-2, ..., l-C$, due to multiple (dense) skip connections. Calculating FCE is not trivial, since we FMotifs from different levels are inside the set of causes.

\subsection{Model architectural hyperparameters}
We use the following notation for a convolution operation window and its respective kernel dimensions
\[
\convdescr{width}{height}{stride}{c_{in}}{c_{out}}
\] for a window moving with $stride$ of specified $width$ and $height$, $c_{in}$ the feature depth (image channels) of the input, $c_{out}$ the count of neurons in a hypercolumn or equally the output feature depth.

All CNNs for tiny color images have a stem that is a single convolutional layer 

\[\convdescr{3}{3}{1}{3}{c^{(stem)}_{out}} \quad \text{while other layers are} \quad \convdescr{3}{3}{s}{c^{(k)}_{in}}{c^{(k)}_{out}}\]

a stride of $s=2$ in convolutional layers performs spatial downsampling in ResNets while in DenseNet this is done with a max pooling operation of stride 2.

\begin{table}[!ht]
    \centering
    \caption{Architectures and feature setups. Residual networks have an \textit{R} prefix and DenseNet \textit{D}. K=Layers, B=Bottleneck, C=Compression $50\%$, k=Feature expansion size.}
    \begin{tabular}{llcccccc}
        \toprule    
            Id
          & Layers
          & $c^{(stem)}_{out}$ & Block1 & Block2 & Block3 & Block4 & Block5  \\
        \midrule    
R1 & K=18 & 16 & 16 & 16 & 32 & 64 & ~ \\
R8 & K=18 & 8 & 8 & 8 & 8 & 8 & ~ \\
R16 & K=18 & 16 & 16 & 16 & 16 & 16 & ~ \\               
R24 & K=18 & 24 & 24 & 24 & 24 & 24 & ~ \\               
R16 & K=18 & 16 & 16 & 16 & 16 & 16 & ~ \\     
R32 & K=18 & 32 & 32 & 32 & 32 & 32 & ~ \\       
R48 & K=18 & 48 & 48 & 48 & 48 & 48 & ~ \\             
R64 & K=18 & 64 & 64 & 64 & 64 & 64 & ~ \\               
R72 & K=18 & 72 & 72 & 72 & 72 & 72 & ~ \\     
R80 & K=18 & 80 & 80 & 80 & 80 & 80 & ~ \\     
R1  & K=20 & 16 & 16 & 16 & 32 & 64 & ~ \\
R8 & K=22 & 8 & 8 & 8 & 8 & 8 & 8 \\
R16 & K=22 & 16 & 16 & 16 & 16 & 16 & 16 \\               
R32 & K=22 & 32 & 32 & 32 & 32 & 32 & 32  \\       
R48 & K=22 & 48 & 48 & 48 & 48 & 48 & 48 \\             
R64 & K=22 & 64 & 64 & 64 & 64 & 64 & 64 \\               
R8 & K=26 & 8 & 8 & 8 & 8 & 8 & ~ \\
R16 & K=26 & 16 & 16 & 16 & 16 & 16 & ~ \\               
R32 & K=26 & 32 & 32 & 32 & 32 & 32 & ~ \\       
R64 & K=26 & 64 & 64 & 64 & 64 & 64 & ~ \\    
D40 & K=100 & 24 & k=12 & k=12 & k=12 & ~ & ~ \\
D100BC & K=100 & 24 & k=12 & f=12 & k=12 & ~ & ~ \\
        \bottomrule         
    \end{tabular}
\end{table}

\subsection{Supervised training hyperparameters}

\paragraph{MNIST:}
We used SGD with an initial learning rate $lr_{@0}=0.02$ and $0.9$ momentum, and we change learning rate to $lr_{@15}=0.01$, $lr_{@30}=0.005$, $lr_{@40}=0.002$, $lr_{@50}=0.001$ until the terminal epoch 60.

\paragraph{FMNIST, KMNIST, OMNIST:}
A slightly different scheduling compared to MNIST is used for the rest of tiny image grayscale datasets, keeping the same initial learning rate and momentum. The SGD starts with a learning rate $lr_{@0}=0.02$, and changes to $lr_{@15}=0.01$, $lr_{@35}=0.005$, $lr_{@45}=0.002$, $lr_{@55}=0.001$, $lr_{@65}=0.0005$ until the terminal epoch 75.

\paragraph{CIFAR10:} We are using the setup in \cite{he_deep_2016} for all training hyperparameters. SGD with $0.9$ momentum starts with a $lr_{@0}=0.1$ and this is divided by 10 at epochs 82, 123 having 391 steps/epoch, until the terminal epoch 164, which is slightly above 64000 steps that are described in the original paper. The same training setup is also used for DenseNets.

\subsection{Software implementation}
This work is enabled by proof-of-concept software implementation that is based on TensorFlow/Keras and popular Python packages numpy, matplotlib, pandas. It includes the first version of the explainability framework that is working on a single process, parts of which are not yet enabled for GPU acceleration. All datasets, models and processes that are not supported by the frameworks have been implemented from scratch. The source code includes a reusable library for Machine Learning, that will be released in the future as a standalone package and registered in the PyPI repository.  The software is available for use by researchers on the primary author's GitHub repository \textbf{**redacted**}, along with its documentation on how to setup and use it. 

\paragraph{Enabling reproducibility of training process and final state}
 Several aspects of experiment reproducibility were taken into account for training models, to have the same intermediate states at the end of each epoch and result to same performance metrics.
 
 The non-deterministic behaviour of 2D convolution algorithms for the GPU was disabled and cuDNN was forced to use the FFT algorithm for convolution. For TensorFlow, op determinism was enabled, while TensorFlow32 and mixed precision were disabled.  These provisions will be release as part of the upcoming Python package.

Nevertheless, training on different infrastructures resulted in different outcomes for the baseline ResNet models. This is an important finding to be reported: Reproducibility of training states and final values of model parameters is not assured when different combinations of hardware, versions of CUDA+cuDNN middleware and versions of computational framework software are used. Reasons behind this should be further understood.

\subsection{Machine Learning infrastructure}
\subsubsection{Requirements}
The available compute infrastructure for Machine Learning allowed us to run multiple single model training processes for the needs of our work. Our experiments are done with small models, so a nominal requirement for them to run would be a single CUDA-compatible GPU with 10GB. The training time per epoch was estimated for each model and was records in the experiment's workspace, that is a file structure that our software library creates.

\subsubsection{Used resources}
The final experiments were performed on 14 NVIDIA A16 GPUs, each one with 16GB, placed in two systems that run Ubuntu 22.04 operating system. For the preliminary experiments 2 NVIDIA RTX3060 GPUs with 12GB were available on a Windows Server 2022 DataCenter system.

\begin{itemize}
\item The middleware used for NVIDIA A16 was CUDA 12.2 + cuDNN 8.9.6 and the software library was TensorFlow 2.15.1. The complete version details for the Ubuntu system infrastructure are recorded in each experiment log file.
\item The middleware used for RTX3060 was CUDA 11.7 + cuDNN DLL 6.14.11.640 and the software library was TensorFlow 2.9.3. The complete version details for the Windows system infrastructure are recorded in each experiment log file.
\end{itemize}

\FloatBarrier
\clearpage
\section{Appendix / Complete set of model evaluation metrics}
\label{sec:appD}

\subsection{Experiments on MNIST}
\paragraph{Initial comparison with same weight decay}

Our first series of experiments were done on MNIST. We investigated performance for a fixed size of 18 layers by increasing the feature count. Starting from 8 we increase $+8$ until 32, then $+16$ until 64, then $+8$ until 80. Additional we used the inverse pyramid features of ResNet for CIFAR10, following the setup in \cite{he_deep_2016}.
The rest of the experiments combined higher number of layers with different features setups. Both ResNets and R-ExplaiNets were trained by keeping the same training hyperparameters, with weight decay $\lambda=0.001$.

\setlength{\tabcolsep}{4pt} 
\begin{table}[!ht]
    \centering
    \caption{Classification accuracy and relative model efficiency for group trained on MNIST-1.}
    \begin{tabular}{lccccccccc}
    \toprule
Architecture &
Id &
Layers &
Features &
MP &
Folds &
Average &
Min &
Max &
RME \\
    \midrule
R-ExplaiNet & R01 & 18 & 16, 16, 16, 32, 64 & 0.18 & 10 & 0.367\ci{0.027} & 0.280 & 0.430 & 0.87 \\
R-ExplaiNet & R08 & 18 & f=8 & 0.01 & 10 & 0.560\ci{0.036} & 0.460 & 0.650 & 0.00 \\
R-ExplaiNet & R16 & 18 & f=16 & 0.04 & 10 & 0.374\ci{0.034} & 0.260 & 0.450 & 1.76 \\
R-ExplaiNet & R24 & 18 & f=24 & 0.09 & 10 & 0.324\ci{0.022} & 0.270 & 0.380 & 2.01 \\
R-ExplaiNet & R32 & 18 & f=32 & 0.15 & 10 & 0.301\ci{0.021} & 0.250 & 0.340 & 1.87 \\
R-ExplaiNet & R48 & 18 & f=48 & 0.34 & 10 & 0.296\ci{0.016} & 0.250 & 0.330 & 1.31 \\
R-ExplaiNet & R64 & 18 & f=64 & 0.60 & 10 & 0.289\ci{0.019} & 0.220 & 0.330 & 1.04 \\
R-ExplaiNet & R72 & 18 & f=72 & 0.75 & 10 & 0.285\ci{0.008} & 0.270 & 0.310 & 0.96 \\
R-ExplaiNet & R80 & 18 & f=80 & 0.93 & 10 & 0.276\ci{0.018} & 0.230 & 0.330 & 0.93 \\
R-ExplaiNet & R08 & 22 & f=8 & 0.01 & 10 & 0.519\ci{0.040} & 0.390 & 0.600 & 0.14 \\
R-ExplaiNet & R16 & 22 & f=16 & 0.05 & 10 & 0.328\ci{0.030} & 0.230 & 0.400 & 2.58 \\
R-ExplaiNet & R32 & 22 & f=32 & 0.19 & 10 & 0.290\ci{0.012} & 0.260 & 0.320 & 1.85 \\
R-ExplaiNet & R64 & 22 & f=64 & 0.74 & 10 & 0.268\ci{0.023} & \textbf{0.200} & 0.320 & 1.11 \\
R-ExplaiNet & R08 & 26 & f=8 & 0.01 & 10 & 0.499\ci{0.033} & 0.410 & 0.550 & 0.27 \\
R-ExplaiNet & R16 & 26 & f=16 & 0.06 & 10 & 0.336\ci{0.030} & 0.270 & 0.410 & 2.18 \\
R-ExplaiNet & R32 & 26 & f=32 & 0.23 & 10 & 0.293\ci{0.025} & 0.230 & 0.380 & 1.64 \\
R-ExplaiNet & R64 & 26 & f=64 & 0.89 & 10 & 0.293\ci{0.020} & 0.240 & 0.340 & 0.82 \\
ResNet & R01 & 18 & 16, 16, 16, 32, 64 & 0.18 & 10 & 0.343\ci{0.018} & 0.290 & 0.380 & 1.13 \\
ResNet & R08 & 18 & f=8 & 0.01 & 10 & 0.562\ci{0.027} & 0.480 & 0.630 & 0.00 \\
ResNet & R16 & 18 & f=16 & 0.04 & 10 & 0.387\ci{0.019} & 0.330 & 0.440 & 1.50 \\
ResNet & R24 & 18 & f=24 & 0.09 & 10 & 0.343\ci{0.023} & 0.270 & 0.390 & 1.66 \\
ResNet & R32 & 18 & f=32 & 0.15 & 10 & 0.323\ci{0.018} & 0.270 & 0.370 & 1.53 \\
ResNet & R48 & 18 & f=48 & 0.34 & 10 & 0.301\ci{0.016} & 0.260 & 0.350 & 1.25 \\
ResNet & R64 & 18 & f=64 & 0.60 & 10 & 0.287\ci{0.013} & 0.250 & 0.320 & 1.06 \\
ResNet & R72 & 18 & f=72 & 0.75 & 10 & 0.287\ci{0.015} & 0.260 & 0.330 & 0.95 \\
ResNet & R80 & 18 & f=80 & 0.93 & 10 & 0.282\ci{0.022} & 0.210 & 0.320 & 0.89 \\
ResNet & R08 & 22 & f=8 & 0.01 & 10 & 0.462\ci{0.022} & 0.420 & 0.540 & 0.79 \\
ResNet & R16 & 22 & f=16 & 0.05 & 10 & 0.344\ci{0.020} & 0.290 & 0.390 & 2.19 \\
ResNet & R32 & 22 & f=32 & 0.19 & 10 & 0.275\ci{0.016} & 0.240 & 0.330 & 2.09 \\
ResNet & R64 & 22 & f=64 & 0.74 & 10 & \textbf{0.263}\ci{0.015} & 0.240 & 0.310 & 1.16 \\
ResNet & R08 & 26 & f=8 & 0.01 & 10 & 0.488\ci{0.030} & 0.410 & 0.570 & 0.38 \\
ResNet & R16 & 26 & f=16 & 0.06 & 10 & 0.328\ci{0.027} & 0.260 & 0.400 & \textbf{2.36} \\
ResNet & R32 & 26 & f=32 & 0.23 & 10 & 0.302\ci{0.022} & 0.250 & 0.360 & 1.52 \\
ResNet & R64 & 26 & f=64 & 0.89 & 10 & 0.273\ci{0.015} & 0.210 & 0.300 & 0.98 \\
    \bottomrule
    \end{tabular}
\end{table}

\FloatBarrier
\clearpage
\paragraph{Increased weight decay for training ExplaiNets:}
We trained ResNets with a weight decay of $\lambda=0.001$ and R-ExplaiNets with double amount of regularization using $\lambda=0.002$. All other training conditions were kept the same.

\setlength{\tabcolsep}{4pt} 
\begin{table}[!ht]
    \centering
    \caption{Classification accuracy and relative model efficiency for group trained on MNIST-2}
    \begin{tabular}{lccccccccc}
    \toprule
Architecture &
Id &
Layers &
Features &
MP &
Folds &
Average &
Min &
Max &
RME \\
    \midrule
R-ExplaiNet & R01 & 18 & 16, 16, 16, 32, 64 & 0.18 & 10 & 0.334\ci{0.013} & 0.300 & 0.360 & 1.22 \\
R-ExplaiNet & R08 & 18 & f=8 & 0.01 & 10 & 0.570\ci{0.039} & 0.480 & 0.670 & 0.00 \\
R-ExplaiNet & R16 & 18 & f=16 & 0.04 & 10 & 0.382\ci{0.015} & 0.350 & 0.420 & 1.60 \\
R-ExplaiNet & R24 & 18 & f=24 & 0.09 & 10 & 0.331\ci{0.021} & 0.280 & 0.380 & 1.84 \\
R-ExplaiNet & R32 & 18 & f=32 & 0.15 & 10 & 0.311\ci{0.019} & 0.270 & 0.370 & 1.67 \\
R-ExplaiNet & R48 & 18 & f=48 & 0.34 & 10 & 0.290\ci{0.017} & 0.230 & 0.320 & 1.34 \\
R-ExplaiNet & R64 & 18 & f=64 & 0.60 & 10 & 0.300\ci{0.016} & 0.270 & 0.360 & 0.92 \\
R-ExplaiNet & R72 & 18 & f=72 & 0.75 & 10 & 0.282\ci{0.009} & 0.260 & 0.300 & 0.95 \\
R-ExplaiNet & R80 & 18 & f=80 & 0.93 & 10 & 0.269\ci{0.021} & \textbf{0.210} & 0.330 & 0.95 \\
R-ExplaiNet & R08 & 22 & f=8 & 0.01 & 10 & 0.521\ci{0.047} & 0.340 & 0.600 & 0.16 \\
R-ExplaiNet & R16 & 22 & f=16 & 0.05 & 10 & 0.352\ci{0.015} & 0.320 & 0.390 & 1.99 \\
R-ExplaiNet & R32 & 22 & f=32 & 0.19 & 10 & 0.291\ci{0.021} & 0.220 & 0.340 & 1.77 \\
R-ExplaiNet & R64 & 22 & f=64 & 0.74 & 10 & \textbf{0.258}\ci{0.008} & 0.240 & 0.280 & 1.16 \\
R-ExplaiNet & R08 & 26 & f=8 & 0.01 & 10 & 0.487\ci{0.021} & 0.400 & 0.520 & 0.44 \\
R-ExplaiNet & R16 & 26 & f=16 & 0.06 & 10 & 0.347\ci{0.025} & 0.280 & 0.410 & 1.92 \\
R-ExplaiNet & R32 & 26 & f=32 & 0.23 & 10 & 0.303\ci{0.015} & 0.250 & 0.340 & 1.46 \\
R-ExplaiNet & R64 & 26 & f=64 & 0.89 & 10 & 0.297\ci{0.022} & 0.240 & 0.340 & 0.77 \\
ResNet & R01 & 18 & 16, 16, 16, 32, 64 & 0.18 & 10 & 0.343\ci{0.018} & 0.290 & 0.380 & 1.12 \\
ResNet & R08 & 18 & f=8 & 0.01 & 10 & 0.562\ci{0.027} & 0.480 & 0.630 & 0.00 \\
ResNet & R16 & 18 & f=16 & 0.04 & 10 & 0.387\ci{0.019} & 0.330 & 0.440 & 1.50 \\
ResNet & R24 & 18 & f=24 & 0.09 & 10 & 0.343\ci{0.023} & 0.270 & 0.390 & 1.64 \\
ResNet & R32 & 18 & f=32 & 0.15 & 10 & 0.323\ci{0.018} & 0.270 & 0.370 & 1.49 \\
ResNet & R48 & 18 & f=48 & 0.34 & 10 & 0.301\ci{0.016} & 0.260 & 0.350 & 1.22 \\
ResNet & R64 & 18 & f=64 & 0.60 & 10 & 0.287\ci{0.013} & 0.250 & 0.320 & 1.03 \\
ResNet & R72 & 18 & f=72 & 0.75 & 10 & 0.287\ci{0.015} & 0.260 & 0.330 & 0.91 \\
ResNet & R80 & 18 & f=80 & 0.93 & 10 & 0.282\ci{0.022} & \textbf{0.210} & 0.320 & 0.86 \\
ResNet & R16 & 22 & f=16 & 0.05 & 10 & 0.344\ci{0.020} & 0.290 & 0.390 & 2.16 \\
ResNet & R32 & 22 & f=32 & 0.19 & 10 & 0.275\ci{0.016} & 0.240 & 0.330 & 2.02 \\
ResNet & R64 & 22 & f=64 & 0.74 & 10 & 0.263\ci{0.015} & 0.240 & 0.310 & 1.12 \\
ResNet & R16 & 26 & f=16 & 0.06 & 10 & 0.328\ci{0.027} & 0.260 & 0.400 & \textbf{2.31} \\
ResNet & R32 & 26 & f=32 & 0.23 & 10 & 0.302\ci{0.022} & 0.250 & 0.360 & 1.47 \\
ResNet & R64 & 26 & f=64 & 0.89 & 10 & 0.273\ci{0.015} & \textbf{0.210} & 0.300 & 0.94 \\
    \bottomrule
    \end{tabular}
\end{table}

\FloatBarrier
\clearpage
\paragraph{Using input value clipping with the same weight decay for training ExplaiNets}:
We made a modification in LIL so that the input is clipped in the range $[-6,6]$, following Conjecture 1, that is stated under the formal proof of gradient amplification (see Appendix \ref{sec:appA}). We trained both ResNets- and R-ExplaiNets-C with all training hyperparameters kept the same. The following table compares models with clipping that are noted with the suffix $C$,  with R-ExplaiNet models without clipping trained with $\lambda=0.002$, and ResNets with $\lambda=0.001$.

\setlength{\tabcolsep}{4pt} 
\begin{table}[!ht]
    \centering
    \caption{Classification accuracy and relative model efficiency for group trained on MNIST-3}
    \begin{tabular}{lccccccccc}
    \toprule
Architecture &
Id &
Layers &
Features &
MP &
Folds &
Average &
Min &
Max &
RME \\
    \midrule
R-ExplaiNet-C & R08 & 18 & f=8 & 0.01 & 10 & 0.546\ci{0.042} & 0.460 & 0.670 & 0.04 \\
R-ExplaiNet-C & R16 & 18 & f=16 & 0.04 & 10 & 0.364\ci{0.030} & 0.270 & 0.430 & 1.93 \\
R-ExplaiNet-C & R32 & 18 & f=32 & 0.15 & 10 & 0.311\ci{0.014} & 0.270 & 0.350 & 1.64 \\
R-ExplaiNet-C & R64 & 18 & f=64 & 0.60 & 10 & 0.287\ci{0.018} & 0.240 & 0.330 & 1.01 \\
R-ExplaiNet-C & R08 & 22 & f=8 & 0.01 & 10 & 0.513\ci{0.031} & 0.430 & 0.610 & 0.22 \\
R-ExplaiNet-C & R16 & 22 & f=16 & 0.05 & 10 & 0.346\ci{0.022} & 0.300 & 0.420 & 2.09 \\
R-ExplaiNet-C & R32 & 22 & f=32 & 0.19 & 10 & 0.283\ci{0.025} & 0.220 & 0.350 & 1.87 \\
R-ExplaiNet-C & R64 & 22 & f=64 & 0.74 & 10 & \textbf{0.256}\ci{0.013} & 0.220 & 0.290 & 1.16 \\
R-ExplaiNet-C & R08 & 26 & f=8 & 0.01 & 10 & 0.469\ci{0.031} & 0.360 & 0.530 & 0.66 \\
R-ExplaiNet-C & R16 & 26 & f=16 & 0.06 & 10 & 0.354\ci{0.029} & 0.300 & 0.450 & 1.76 \\
R-ExplaiNet-C & R32 & 26 & f=32 & 0.23 & 10 & 0.294\ci{0.023} & 0.240 & 0.350 & 1.55 \\
R-ExplaiNet-C & R64 & 26 & f=64 & 0.89 & 10 & 0.292\ci{0.018} & 0.240 & 0.340 & 0.79 \\
R-ExplaiNet-C & R08 & 18 & f=8 & 0.01 & 10 & 0.570\ci{0.039} & 0.480 & 0.670 & 0.00 \\
R-ExplaiNet-C & R16 & 18 & f=16 & 0.04 & 10 & 0.382\ci{0.015} & 0.350 & 0.420 & 1.57 \\
R-ExplaiNet-C & R32 & 18 & f=32 & 0.15 & 10 & 0.311\ci{0.019} & 0.270 & 0.370 & 1.64 \\
R-ExplaiNet-C & R64 & 18 & f=64 & 0.60 & 10 & 0.300\ci{0.016} & 0.270 & 0.360 & 0.91 \\
R-ExplaiNet & R08 & 22 & f=8 & 0.01 & 10 & 0.521\ci{0.047} & 0.340 & 0.600 & 0.16 \\
R-ExplaiNet & R16 & 22 & f=16 & 0.05 & 10 & 0.352\ci{0.015} & 0.320 & 0.390 & 1.96 \\
R-ExplaiNet & R32 & 22 & f=32 & 0.19 & 10 & 0.291\ci{0.021} & 0.220 & 0.340 & 1.75 \\
R-ExplaiNet & R64 & 22 & f=64 & 0.74 & 10 & 0.258\ci{0.008} & 0.240 & 0.280 & 1.14 \\
R-ExplaiNet & R08 & 26 & f=8 & 0.01 & 10 & 0.487\ci{0.021} & 0.400 & 0.520 & 0.44 \\
R-ExplaiNet & R16 & 26 & f=16 & 0.06 & 10 & 0.347\ci{0.025} & 0.280 & 0.410 & 1.89 \\
R-ExplaiNet & R32 & 26 & f=32 & 0.23 & 10 & 0.303\ci{0.015} & 0.250 & 0.340 & 1.44 \\
R-ExplaiNet & R64 & 26 & f=64 & 0.89 & 10 & 0.297\ci{0.022} & 0.240 & 0.340 & 0.76 \\
ResNet & R08 & 18 & f=8 & 0.01 & 10 & 0.562\ci{0.027} & 0.480 & 0.630 & 0.00 \\
ResNet & R16 & 18 & f=16 & 0.04 & 10 & 0.387\ci{0.019} & 0.330 & 0.440 & 1.48 \\
ResNet & R32 & 18 & f=32 & 0.15 & 10 & 0.323\ci{0.018} & 0.270 & 0.370 & 1.47 \\
ResNet & R64 & 18 & f=64 & 0.60 & 10 & 0.287\ci{0.013} & 0.250 & 0.320 & 1.01 \\
ResNet & R08 & 22 & f=8 & 0.01 & 10 & 0.462\ci{0.022} & 0.420 & 0.540 & 0.83 \\
ResNet & R16 & 22 & f=16 & 0.05 & 10 & 0.344\ci{0.020} & 0.290 & 0.390 & 2.13 \\
ResNet & R32 & 22 & f=32 & 0.19 & 10 & 0.275\ci{0.016} & 0.240 & 0.330 & 1.99 \\
ResNet & R64 & 22 & f=64 & 0.74 & 10 & 0.263\ci{0.015} & 0.240 & 0.310 & 1.10 \\
ResNet & R08 & 26 & f=8 & 0.01 & 10 & 0.488\ci{0.030} & 0.410 & 0.570 & 0.43 \\
ResNet & R16 & 26 & f=16 & 0.06 & 10 & 0.328\ci{0.027} & 0.260 & 0.400 & \textbf{2.28} \\
ResNet & R32 & 26 & f=32 & 0.23 & 10 & 0.302\ci{0.022} & 0.250 & 0.360 & 1.45 \\
ResNet & R64 & 26 & f=64 & 0.89 & 10 & 0.273\ci{0.015} & \textbf{0.210} & 0.300 & 0.93 \\
    \bottomrule
    \end{tabular}
\end{table}

\FloatBarrier
\clearpage
\paragraph{Using input value clipping with increased weight decay for training ExplaiNets:}

The last set of experiments used clipping in the range $[-6,6]$ and $\lambda=0.002$ for R-ExplaiNets-C and the same training conditions with all experiments on MNIST. In the results it was apparent that more regularization lead to instability and worse performance.

\setlength{\tabcolsep}{4pt} 
\begin{table}[!ht]
    \centering
    \caption{Classification accuracy and relative model efficiency for group trained on MNIST-4}
    \begin{tabular}{lccccccccc}
    \toprule
Architecture &
Id &
Layers &
Features &
MP &
Folds &
Average &
Min &
Max &
RME \\
    \midrule
R-ExplaiNet-C & R08 & 18 & f=8 & 0.01 & 10 & 1.140\ci{0.213} & 0.750 & 1.700 & 1.36 \\
R-ExplaiNet-C & R16 & 18 & f=16 & 0.04 & 10 & 0.479\ci{0.032} & 0.400 & 0.540 & 3.53 \\
R-ExplaiNet-C & R32 & 18 & f=32 & 0.15 & 10 & 0.319\ci{0.014} & 0.290 & 0.350 & 2.36 \\
R-ExplaiNet-C & R64 & 18 & f=64 & 0.60 & 10 & 0.267\ci{0.014} & 0.230 & 0.300 & 1.29 \\
R-ExplaiNet-C & R08 & 22 & f=8 & 0.01 & 10 & 1.752\ci{0.851} & 0.720 & 5.200 & 0.00 \\
R-ExplaiNet-C & R16 & 22 & f=16 & 0.05 & 10 & 0.452\ci{0.023} & 0.370 & 0.510 & 3.32 \\
R-ExplaiNet-C & R32 & 22 & f=32 & 0.19 & 10 & 0.296\ci{0.012} & 0.280 & 0.340 & 2.19 \\
R-ExplaiNet-C & R64 & 22 & f=64 & 0.74 & 10 & 0.271\ci{0.021} & 0.220 & 0.320 & 1.14 \\
R-ExplaiNet-C & R08 & 26 & f=8 & 0.01 & 10 & 1.270\ci{0.229} & 0.850 & 1.910 & 0.67 \\
R-ExplaiNet-C & R16 & 26 & f=16 & 0.06 & 10 & 0.430\ci{0.034} & 0.330 & 0.510 & 3.16 \\
R-ExplaiNet-C & R32 & 26 & f=32 & 0.23 & 10 & 0.333\ci{0.016} & 0.280 & 0.370 & 1.88 \\
R-ExplaiNet-C & R64 & 26 & f=64 & 0.89 & 10 & 0.291\ci{0.015} & 0.260 & 0.330 & 1.01 \\
ResNet & R08 & 18 & f=8 & 0.01 & 10 & 0.562\ci{0.027} & 0.480 & 0.630 & 5.94 \\
ResNet & R16 & 18 & f=16 & 0.04 & 10 & 0.387\ci{0.019} & 0.330 & 0.440 & 4.16 \\
ResNet & R32 & 18 & f=32 & 0.15 & 10 & 0.323\ci{0.018} & 0.270 & 0.370 & 2.34 \\
ResNet & R64 & 18 & f=64 & 0.60 & 10 & 0.287\ci{0.013} & 0.250 & 0.320 & 1.25 \\
ResNet & R08 & 22 & f=8 & 0.01 & 10 & 0.462\ci{0.022} & 0.420 & 0.540 & \textbf{6.42} \\
ResNet & R16 & 22 & f=16 & 0.05 & 10 & 0.344\ci{0.020} & 0.290 & 0.390 & 4.01 \\
ResNet & R32 & 22 & f=32 & 0.19 & 10 & 0.275\ci{0.016} & 0.240 & 0.330 & 2.26 \\
ResNet & R64 & 22 & f=64 & 0.74 & 10 & \textbf{0.263}\ci{0.015} & 0.240 & 0.310 & 1.16 \\
ResNet & R08 & 26 & f=8 & 0.01 & 10 & 0.488\ci{0.030} & 0.410 & 0.570 & 5.59 \\
ResNet & R16 & 26 & f=16 & 0.06 & 10 & 0.328\ci{0.027} & 0.260 & 0.400 & 3.76 \\
ResNet & R32 & 26 & f=32 & 0.23 & 10 & 0.302\ci{0.022} & 0.250 & 0.360 & 1.98 \\
ResNet & R64 & 26 & f=64 & 0.89 & 10 & 0.273\ci{0.015} & \textbf{0.210} & 0.300 & 1.04 \\
    \bottomrule
    \end{tabular}
\end{table}

\FloatBarrier
\clearpage
\subsection{Experiments on Fashion MNIST}

In the table below $Id$ identifies the architectural hyperparameter setup for the the depth of layers that corresponds to features per block, where $f=$ stands for the same count of features in all layers. Model size is reported in  millions of parameters (\modelsize{}).

\setlength{\tabcolsep}{4pt} 
\begin{table}[!ht]
    \centering
    \caption{Classification accuracy and relative model efficiency for group trained on FMNIST}
    \begin{tabular}{lccccccccc}
    \toprule
Architecture &
Id &
Layers &
Features &
MP &
Folds &
Average &
Min &
Max &
RME \\
    \midrule
R-ExplaiNet & R01 & 18 & 16, 16, 16, 32, 64 & 0.18 & 10 & 92.02\ci{0.14} & 91.67 & 92.28 & 0.67 \\
R-ExplaiNet & R08 & 18 & f=8 & 0.01 & 10 & 90.71\ci{0.13} & 90.39 & 91.09 & 0.01 \\
R-ExplaiNet & R16 & 18 & f=16 & 0.04 & 10 & 91.53\ci{0.14} & 91.15 & 91.81 & 0.58 \\
R-ExplaiNet & R32 & 18 & f=32 & 0.15 & 10 & 92.41\ci{0.10} & 92.20 & 92.63 & \textbf{1.28} \\
R-ExplaiNet & R64 & 18 & f=64 & 0.60 & 10 & 92.60\ci{0.10} & 92.28 & 92.85 & 0.82 \\
R-ExplaiNet & R01 & 20 & 16, 16, 32, 64 & 0.27 & 10 & 92.59\ci{0.19} & 92.01 & 92.95 & 1.19 \\
R-ExplaiNet & R48 & 22 & f=48 & 0.42 & 10 & 92.86\ci{0.13} & 92.54 & 93.12 & 1.30 \\
R-ExplaiNet & R48 & 26 & f=48 & 0.50 & 10 & 92.91\ci{0.15} & 92.40 & 93.21 & 1.26 \\
R-ExplaiNet & R64 & 26 & f=64 & 0.89 & 10 & \textbf{93.03}\ci{0.12} & 92.70 & \textbf{93.45} & 1.06 \\
ResNet & R01 & 18 & 16, 16, 16, 32, 64 & 0.18 & 10 & 91.89\ci{0.14} & 91.52 & 92.29 & 0.54 \\
ResNet & R08 & 18 & f=8 & 0.01 & 10 & 90.62\ci{0.11} & 90.29 & 90.85 & 0.00 \\
ResNet & R16 & 18 & f=16 & 0.04 & 10 & 91.28\ci{0.13} & 90.86 & 91.56 & 0.29 \\
ResNet & R32 & 18 & f=32 & 0.15 & 10 & 92.09\ci{0.09} & 91.88 & 92.40 & 0.82 \\
ResNet & R64 & 18 & f=64 & 0.60 & 10 & 92.54\ci{0.10} & 92.28 & 92.76 & 0.76 \\
ResNet & R01 & 20 & 16, 16, 32, 64 & 0.27 & 10 & 92.42\ci{0.12} & 92.14 & 92.64 & 0.97 \\
ResNet & R48 & 22 & f=48 & 0.42 & 10 & 92.69\ci{0.11} & 92.34 & 92.89 & 1.08 \\
ResNet & R48 & 26 & f=48 & 0.50 & 10 & 92.84\ci{0.14} & 92.54 & 93.19 & 1.17 \\
ResNet & R64 & 26 & f=64 & 0.89 & 10 & 92.83\ci{0.13} & 92.40 & 93.15 & 0.86 \\
    \bottomrule
    \end{tabular}
\end{table}

\subsection{Experiments on Kuzushiji MNIST}

\setlength{\tabcolsep}{4pt} 
\begin{table}[!ht]
    \centering
    \caption{Classification accuracy and relative model efficiency for group trained on KMNIST}
    \begin{tabular}{lccccccccc}
    \toprule
Architecture &
Id &
Layers &
Features &
MP &
Folds &
Average &
Min &
Max &
RME \\
    \midrule
R-ExplaiNet & R08 & 18 & f=8 & 0.01 & 10 & 95.33\ci{0.25} & 94.36 & 95.86 & 0.00 \\
R-ExplaiNet & R16 & 18 & f=16 & 0.04 & 10 & 97.63\ci{0.08} & 97.38 & 97.80 & \textbf{2.19} \\
R-ExplaiNet & R32 & 18 & f=32 & 0.15 & 10 & 98.39\ci{0.07} & 98.22 & 98.62 & 2.10 \\
R-ExplaiNet & R64 & 18 & f=64 & 0.60 & 10 & 98.56\ci{0.04} & 98.43 & 98.62 & 1.20 \\
R-ExplaiNet & R01 & 20 & 16, 16, 32, 64 & 0.27 & 10 & 98.62\ci{0.06} & 98.42 & 98.75 & 1.87 \\
R-ExplaiNet & R48 & 22 & f=48 & 0.42 & 10 & 98.59\ci{0.05} & 98.43 & 98.71 & 1.46 \\
R-ExplaiNet & R48 & 26 & f=48 & 0.50 & 10 & 98.53\ci{0.06} & 98.41 & 98.75 & 1.29 \\
R-ExplaiNet & R64 & 26 & f=64 & 0.89 & 10 & \textbf{98.66}\ci{0.05} & 98.53 & \textbf{98.78} & 1.06 \\
ResNet & R08 & 18 & f=8 & 0.01 & 10 & 95.30\ci{0.14} & 94.88 & 95.63 & 0.00 \\
ResNet & R16 & 18 & f=16 & 0.04 & 10 & 97.50\ci{0.09} & 97.29 & 97.73 & 1.91 \\
ResNet & R32 & 18 & f=32 & 0.15 & 10 & 98.16\ci{0.10} & 98.01 & 98.46 & 1.76 \\
ResNet & R64 & 18 & f=64 & 0.60 & 10 & 98.39\ci{0.06} & 98.22 & 98.56 & 1.06 \\
ResNet & R01 & 20 & 16, 16, 32, 64 & 0.27 & 10 & 98.43\ci{0.08} & 98.23 & 98.67 & 1.63 \\
ResNet & R48 & 22 & f=48 & 0.42 & 10 & 98.44\ci{0.05} & 98.34 & 98.58 & 1.32 \\
ResNet & R48 & 26 & f=48 & 0.50 & 10 & 98.42\ci{0.06} & 98.24 & 98.52 & 1.18 \\
ResNet & R64 & 26 & f=64 & 0.89 & 10 & 98.53\ci{0.06} & 98.34 & 98.66 & 0.97 \\
    \bottomrule
    \end{tabular}
\end{table}

\FloatBarrier
\clearpage
\subsection{Experiments on Oracle MNIST}

\setlength{\tabcolsep}{4pt} 
\begin{table}[!ht]
    \centering
    \caption{Classification accuracy and relative model efficiency for group trained on OMNIST}
    \begin{tabular}{lccccccccc}
    \toprule
Architecture &
Id &
Layers &
Features &
MP &
Folds &
Average &
Min &
Max &
RME \\
    \midrule
R-ExplaiNet & R08 & 18 & f=8 & 0.01 & 10 & 93.35\ci{0.17} & 92.83 & 93.67 & 0.03 \\
R-ExplaiNet & R16 & 18 & f=16 & 0.04 & 10 & 95.35\ci{0.09} & 95.17 & 95.60 & \textbf{1.74} \\
R-ExplaiNet & R32 & 18 & f=32 & 0.15 & 10 & 96.13\ci{0.13} & 95.90 & 96.47 & 1.74 \\
R-ExplaiNet & R64 & 18 & f=64 & 0.60 & 10 & 96.49\ci{0.17} & 96.10 & 96.87 & 1.13 \\
R-ExplaiNet & R01 & 20 & 16, 16, 32, 64 & 0.27 & 10 & 96.33\ci{0.14} & 95.90 & 96.67 & 1.50 \\
R-ExplaiNet & R48 & 22 & f=48 & 0.42 & 10 & 96.53\ci{0.13} & 96.13 & 96.83 & 1.40 \\
R-ExplaiNet & R48 & 26 & f=48 & 0.50 & 20 & 96.65\ci{0.07} & 96.27 & 96.93 & 1.38 \\
R-ExplaiNet & R64 & 26 & f=64 & 0.89 & 10 & \textbf{96.68}\ci{0.11} & 96.43 & \textbf{96.93} & 1.06 \\
ResNet & R08 & 18 & f=8 & 0.01 & 10 & 93.13\ci{0.30} & 92.17 & 93.73 & 0.00 \\
ResNet & R16 & 18 & f=16 & 0.04 & 10 & 95.21\ci{0.18} & 94.83 & 95.87 & 1.50 \\
ResNet & R32 & 18 & f=32 & 0.15 & 10 & 96.05\ci{0.18} & 95.67 & 96.53 & 1.63 \\
ResNet & R64 & 18 & f=64 & 0.60 & 10 & 96.39\ci{0.15} & 95.93 & 96.77 & 1.06 \\
ResNet & R01 & 20 & 16, 16, 32, 64 & 0.27 & 10 & 96.34\ci{0.12} & 96.03 & 96.63 & 1.51 \\
ResNet & R48 & 22 & f=48 & 0.42 & 10 & 96.39\ci{0.16} & 96.07 & 96.83 & 1.26 \\
ResNet & R48 & 26 & f=48 & 0.50 & 10 & 96.44\ci{0.17} & 95.97 & 96.83 & 1.19 \\
ResNet & R64 & 26 & f=64 & 0.89 & 10 & 96.56\ci{0.12} & 96.33 & 96.93 & 0.97 \\
    \bottomrule
    \end{tabular}
\end{table}

\subsection{Experiments on CIFAR10}

For CIFAR10 we have two different architectures with skip connection where $Id$ identifies the architectural hyperparameter setup for the the depth of layers. For ResNets we have features per block, where $f=$ stands for the same count of features in all layers. For DenseNets we have the feature expansion count $k=$, where $B$ stands for the use of bottleneck $1\times1$ convolutions and $C$ for $50\%$ compression of features in transition layers. Model size is reported in  (\modelsize{}).

\setlength{\tabcolsep}{4pt} 
\begin{table}[!ht]
    \centering
    \caption{Classification accuracy and relative model efficiency for group trained on CIFAR10}
    \begin{tabular}{lccccccccc}
    \toprule
Architecture &
Id &
Layers &
Features &
MP &
Folds &
Average &
Min &
Max &
RME \\
    \midrule
D-ExplaiNet & BC & 100 & k=12 & 0.79 & 5 & \textbf{94.60}\ci{0.16} & 94.35 & \textbf{94.80} & 1.12 \\
DenseNet & BC & 100 & k=12 & 0.79 & 5 & 94.38\ci{0.18} & 94.06 & 94.58 & 1.09 \\
D-ExplaiNet & ~ & 40 & k=12 & 1.08 & 5 & 93.80\ci{0.08} & 93.68 & 93.90 & 0.86 \\
DenseNet & ~ & 40 & k=12 & 1.08 & 5 & 93.41\ci{0.15} & 93.15 & 93.58 & 0.81 \\
R-ExplaiNet & R01 & 18 & 16, 16, 16, 32, 64 & 0.18 & 10 & 89.74\ci{0.11} & 89.48 & 90.04 & 1.04 \\
R-ExplaiNet & R08 & 18 & f=8 & 0.01 & 10 & 78.49\ci{0.09} & 78.20 & 78.70 & 0.00 \\
R-ExplaiNet & R16 & 18 & f=16 & 0.04 & 10 & 87.08\ci{0.11} & 86.82 & 87.34 & 1.26 \\
R-ExplaiNet & R32 & 18 & f=32 & 0.15 & 10 & 91.19\ci{0.11} & 90.80 & 91.42 & \textbf{1.49} \\
R-ExplaiNet & R48 & 18 & f=48 & 0.34 & 10 & 92.61\ci{0.10} & 92.37 & 92.88 & 1.27 \\
R-ExplaiNet & R01 & 20 & 16, 16, 32, 64 & 0.27 & 10 & 91.88\ci{0.13} & 91.52 & 92.28 & 1.26 \\
R-ExplaiNet & R64 & 26 & f=64 & 0.89 & 10 & 93.80\ci{0.12} & 93.54 & 94.15 & 0.94 \\
ResNet & R01 & 18 & 16, 16, 16, 32, 64 & 0.18 & 10 & 89.11\ci{0.11} & 88.93 & 89.46 & 0.91 \\
ResNet & R08 & 18 & f=8 & 0.01 & 10 & 78.16\ci{0.31} & 77.24 & 79.02 & 0.00 \\
ResNet & R16 & 18 & f=16 & 0.04 & 10 & 86.72\ci{0.24} & 86.33 & 87.48 & 1.14 \\
ResNet & R32 & 18 & f=32 & 0.15 & 10 & 90.91\ci{0.13} & 90.62 & 91.32 & 1.42 \\
ResNet & R48 & 18 & f=48 & 0.34 & 10 & 92.16\ci{0.09} & 91.97 & 92.35 & 1.18 \\
ResNet & R01 & 20 & 16, 16, 32, 64 & 0.27 & 10 & 91.59\ci{0.12} & 91.37 & 91.90 & 1.19 \\
ResNet & R64 & 26 & f=64 & 0.90 & 10 & 93.41\ci{0.09} & 93.21 & 93.59 & 0.88 \\
    \bottomrule
    \end{tabular}
\end{table}

\FloatBarrier
\clearpage
\section{Appendix / Feature motif discovery and mosaics of handwritten digits}
\label{sec:appE}

\subsection{Unsupervised feature motif discovery}
We have used YAMDA \cite{quang_yamda_2018} an implementation of EXTREME base on the Torch framework that uses GPU acceleration to discover motifs in genomic sequences. The implementation is available on \href{https://github.com/daquang/YAMDA}{YAMDA GitHub Repository} and we pulled commit \#00e9c9d. Some quality improvements were needed on the implementation provided under the MIT license so that it can function on a sequence set that has short aligned sequences, that is the case our LDF vectors encoded in the quaternary system. 

\paragraph{Sequence data hyperparameters:}
The sequence dataset is converted into the .fasta format using the DNA letter alphabet where $\delta=4$ is the basic data hyperparameter for the algorithm. The batch size is set to 4096, since the maximum size of all LDF vocabularies $V^{(l)}=\{\zldfvector{l}{}(i)\}$ is $|V^{(5)}|=33414$.  In our experiments the size of LDF vectors was $c_{LDF}=4$ for $c_{out}=16$ features, so the length of the quaternary sequence vector is $c_{4}=8$. We use it as our motif width hyperparameter $n_{motif}=8$.

\paragraph{Motif discovery hyperparameters:} Several hyperparameters of YAMDA are kept to their default values, and it is switched to remove a sequence sample from the next epoch, if it contains a discovered motif. We set $n_{half}=2$, the k-mer half-length hyperparameter that is used in the search that seeds the algorithm with the proper initial sequences.  Setting $n_{motif}=8$ makes the algorithm to merely search for motifs in aligned sequences, while it inherently supports motif discovery in any position inside sequences. We set the maximum count of motifs to discover $K_{motif}=96$ and minimum count of occurrences (sites) for a motif in the data to $N_{sites}=200$. 

The algorithm's execution is very fast in our experimental setup, that is promising to scale up to a higher count of CNN features and/or size of LDF vectors.

\subsection{FMotif mosaics of handwritten digit 7.}

\setlength{\belowcaptionskip}{2pt}    
\begin{figure}[!h]
    \centering
    \includegraphics[width=0.7\textwidth]{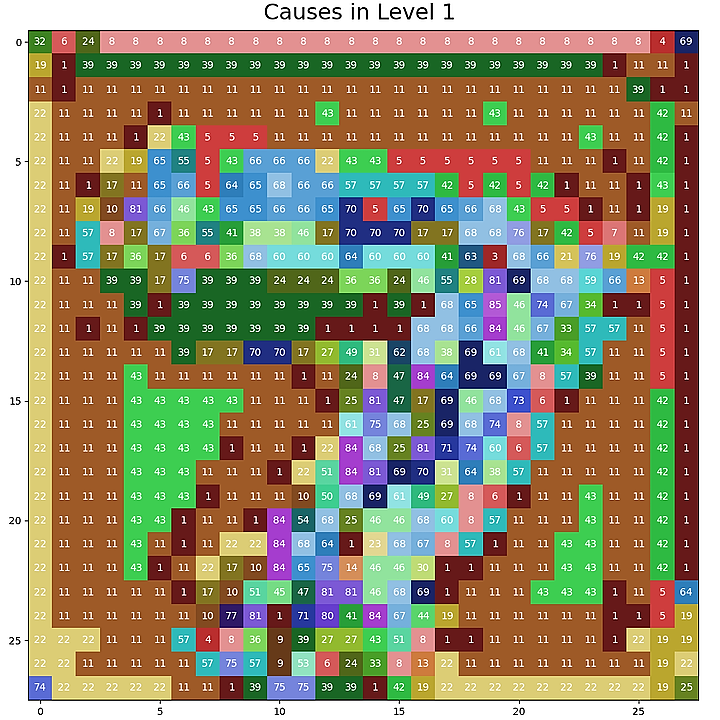}
    \caption{Image representation $\boldsymbol\Psi^{(1)} \in \matrixspacediscrete{28}{28}$ for first sample of the MNIST validation set.}
    \label{fig:L1CausesMosaic}    
\end{figure}

\FloatBarrier
\clearpage

\setlength{\belowcaptionskip}{2pt}
\begin{figure}[!h]
    \centering
    \includegraphics[width=0.7\textwidth]{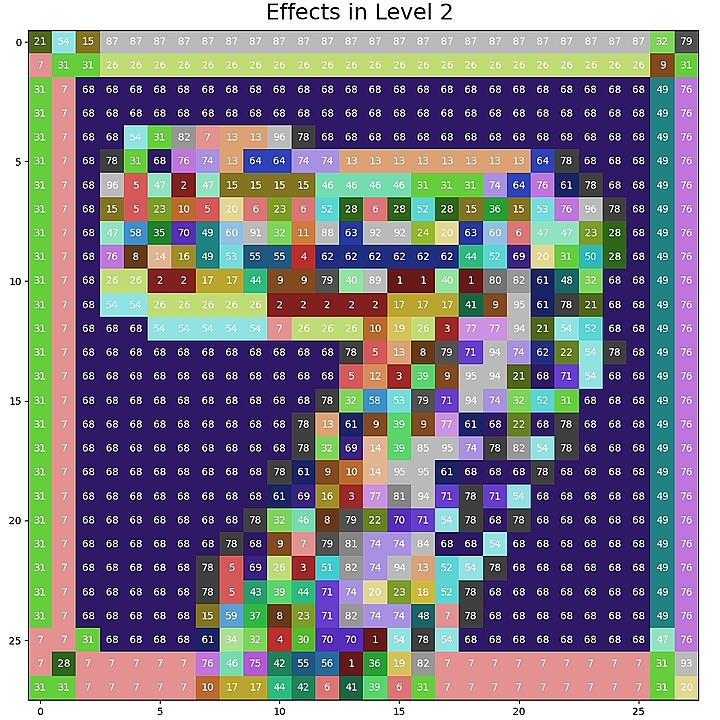}
    \caption{Image representation $\boldsymbol\Psi^{(2)} \in \matrixspacediscrete{28}{28}$ for first sample of the MNIST validation set.}
    \label{fig:L2EffectsMosaic}    
\end{figure}

\subsection{FMotif matching scores for handwritten digits}

\setlength{\belowcaptionskip}{2pt}
\begin{figure}[!h]
    \centering
    \includegraphics[width=0.8\textwidth]{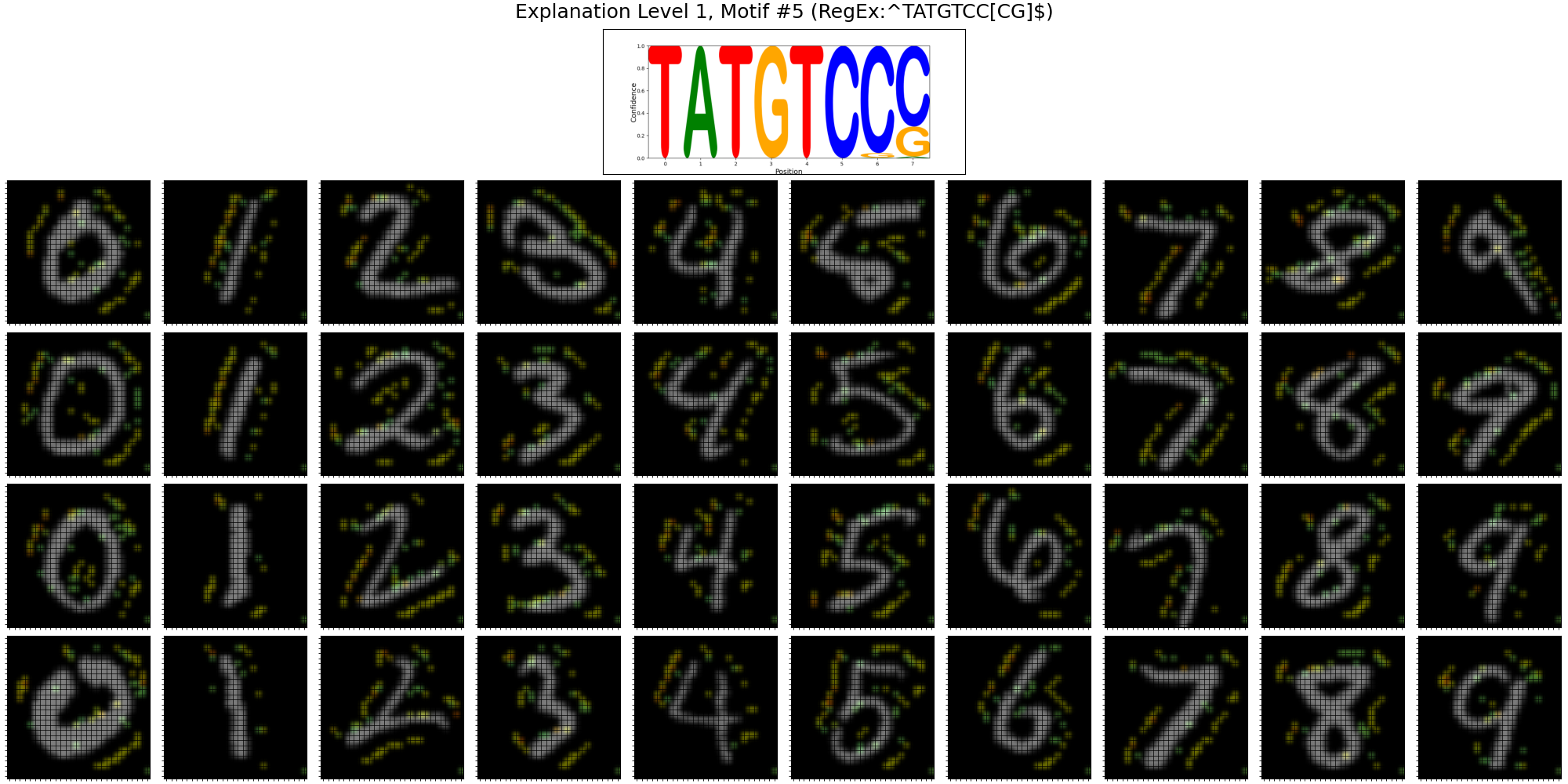}
    \caption{The FMotif 5 that is part of causes in level 1,  seems to present in illumination transition edges. It is visible at row 27 of Figure \ref{fig:L1CausesMosaic} and it is suspected to be a side-effect of zero padding.}
\end{figure}

\FloatBarrier
\clearpage

\setlength{\belowcaptionskip}{2pt}
\begin{figure}[!h]
    \centering
    \includegraphics[width=0.8\textwidth]{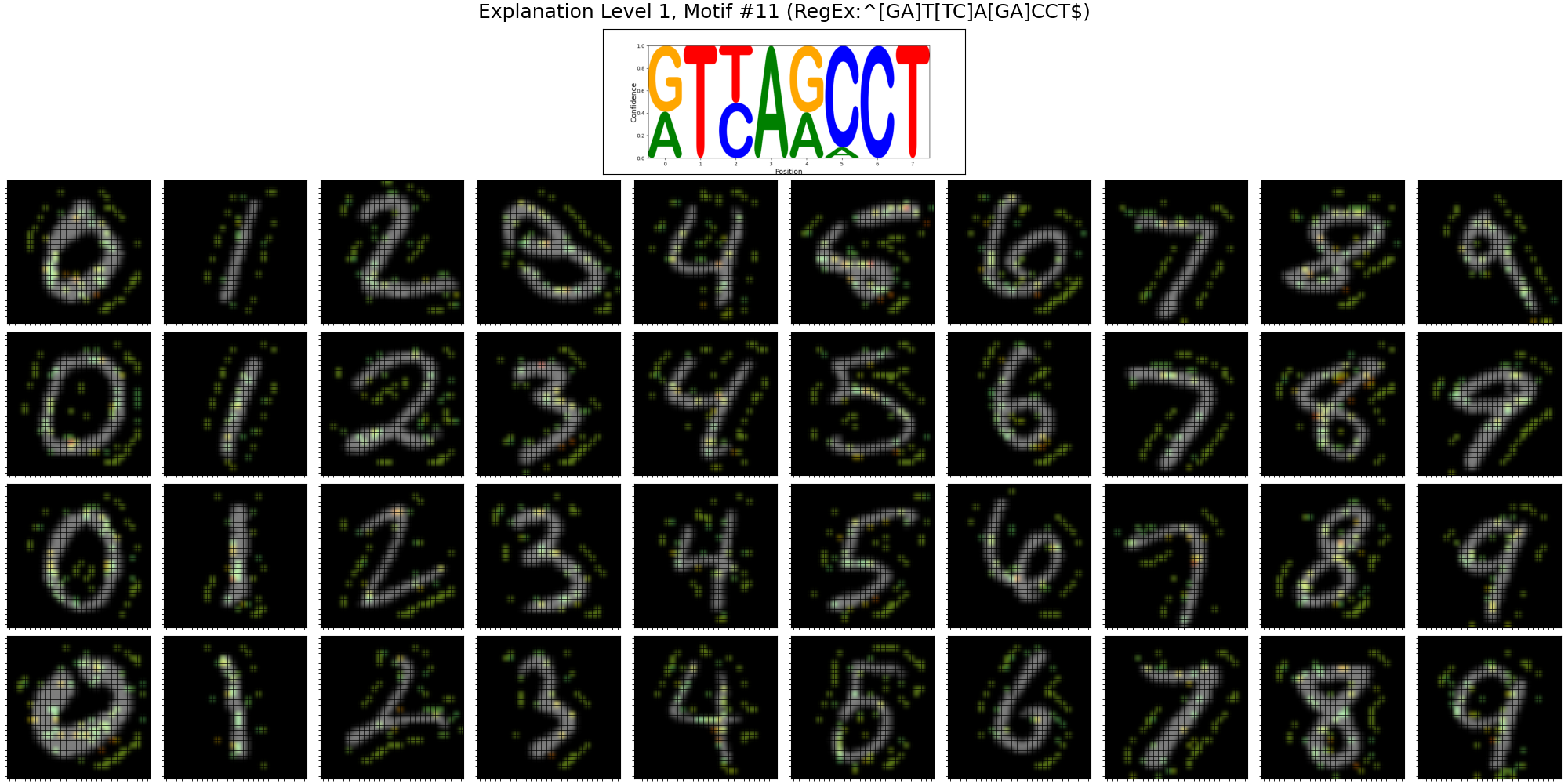}
    \caption{The edge FMotif 11 that is part of causes in level 1.}
\end{figure}

\setlength{\belowcaptionskip}{2pt}
\begin{figure}[!h]
    \centering
    \includegraphics[width=0.8\textwidth]{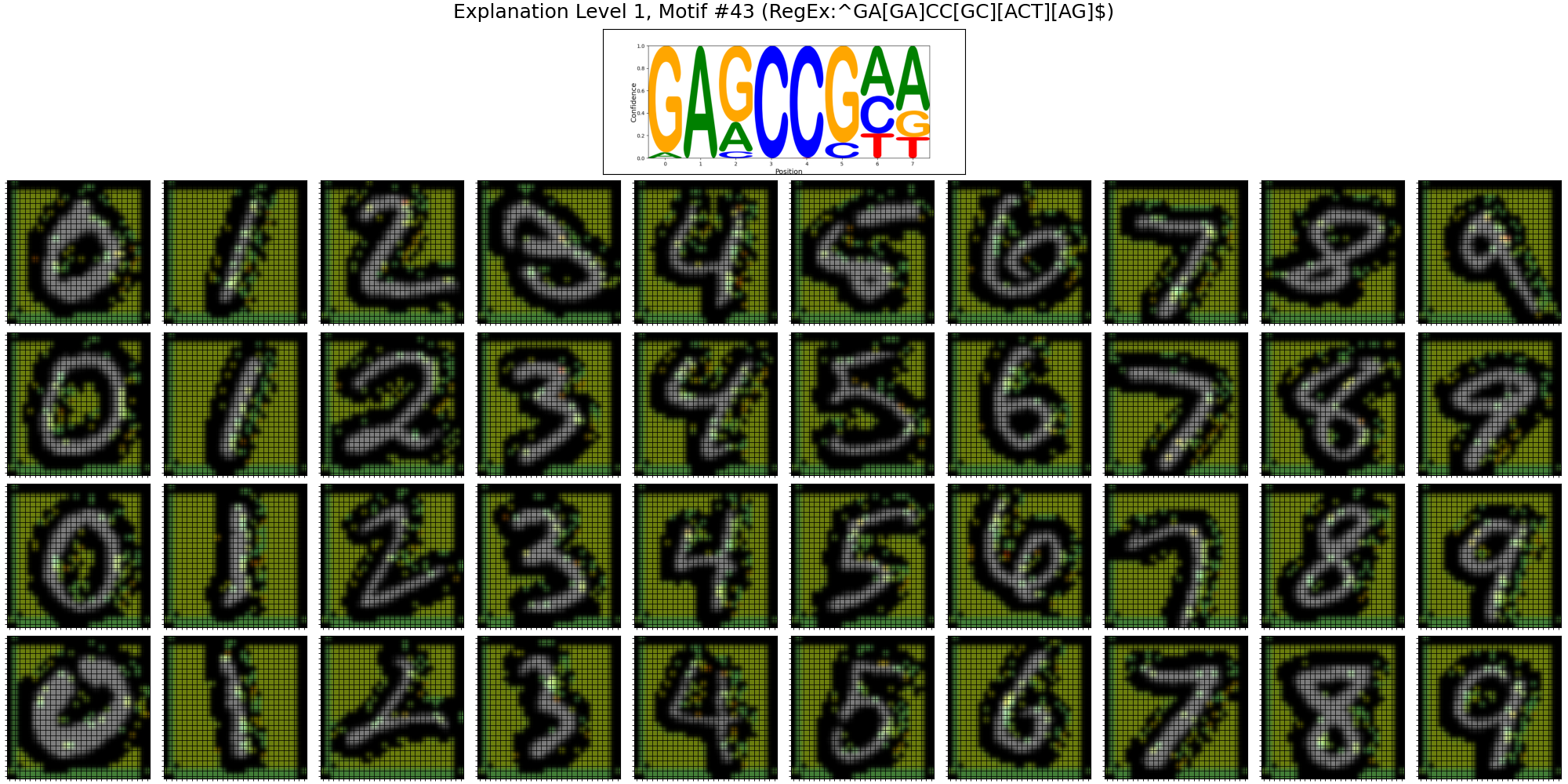}
    \caption{The black background FMotif 43 that is part of causes in level 1.}
\end{figure}

\setlength{\belowcaptionskip}{2pt}
\begin{figure}[!h]
    \centering
    \includegraphics[width=0.8\textwidth]{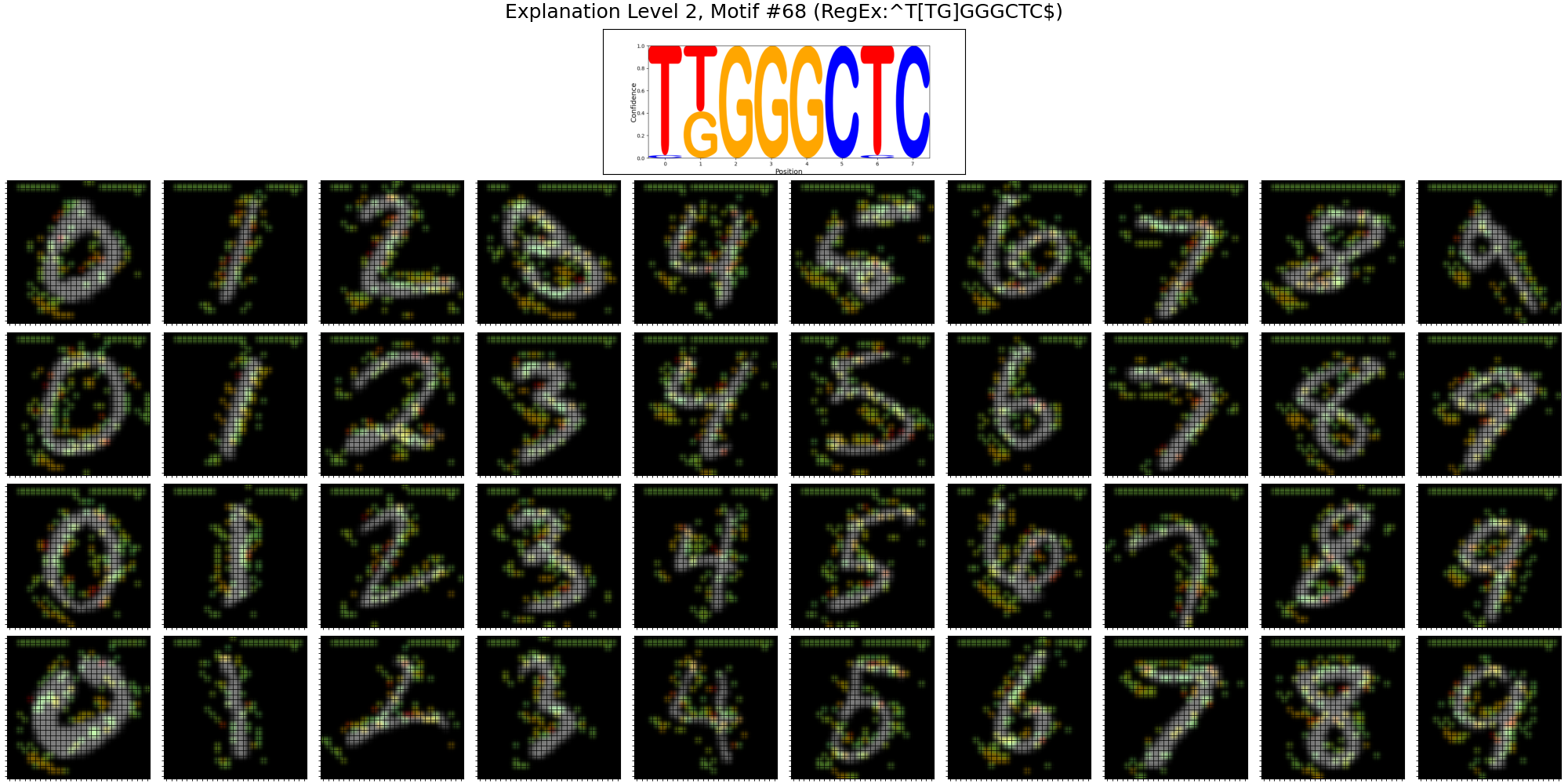}
    \caption{FMotif 68 is observed as an effect in level 2 and dominates many patches as seen in Figure \ref{fig:L2EffectsMosaic}. It can be explained as caused by black background.}
\end{figure}

\FloatBarrier
\clearpage

\setlength{\belowcaptionskip}{2pt}
\begin{figure}[!h]
    \centering
    \includegraphics[width=0.8\textwidth]{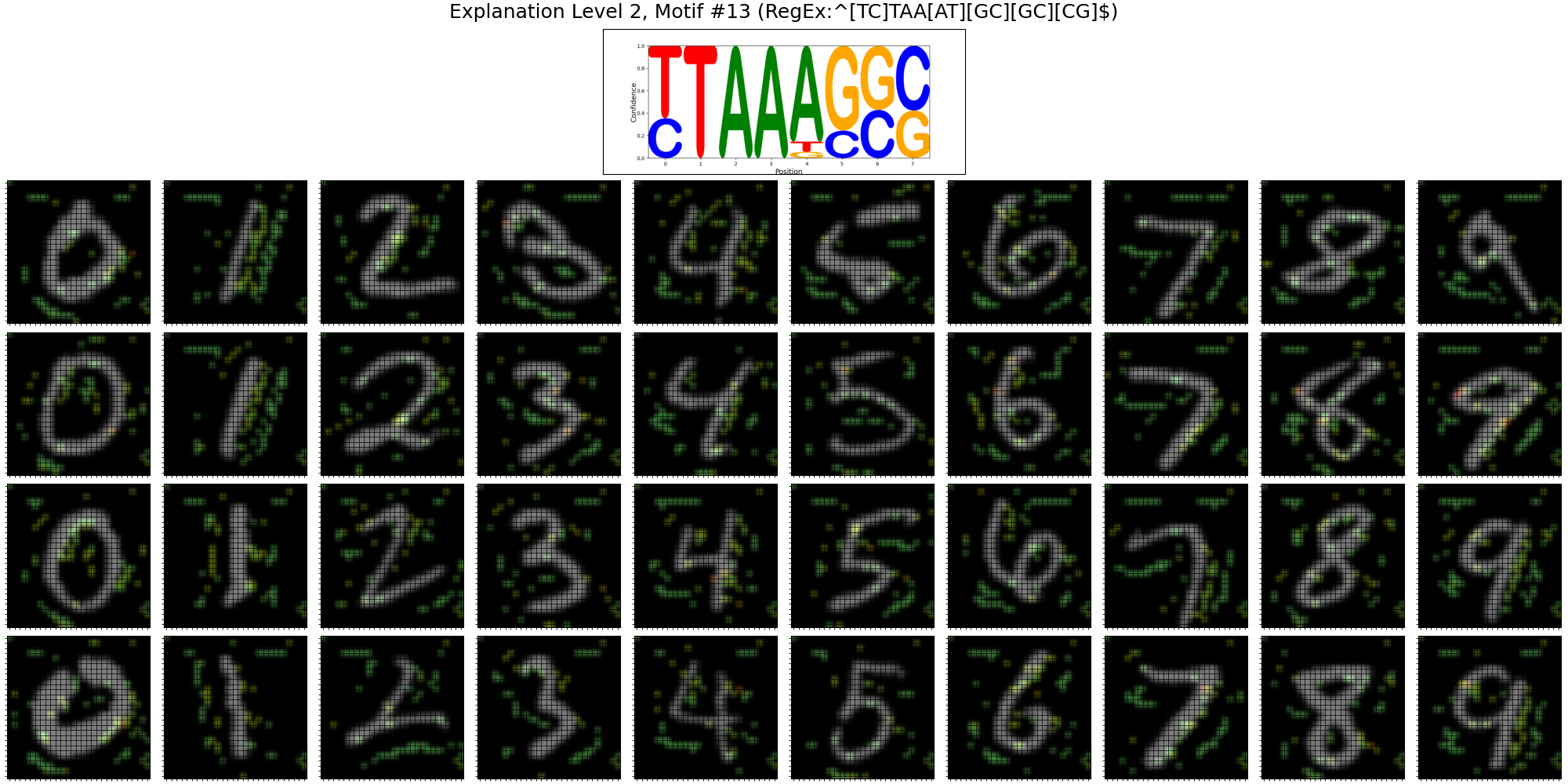}
    \caption{A horizontal edge FMotif 13, observed as effect in level 2.}
\end{figure}

\setlength{\belowcaptionskip}{2pt}
\begin{figure}[!h]
    \centering
    \includegraphics[width=0.8\textwidth]{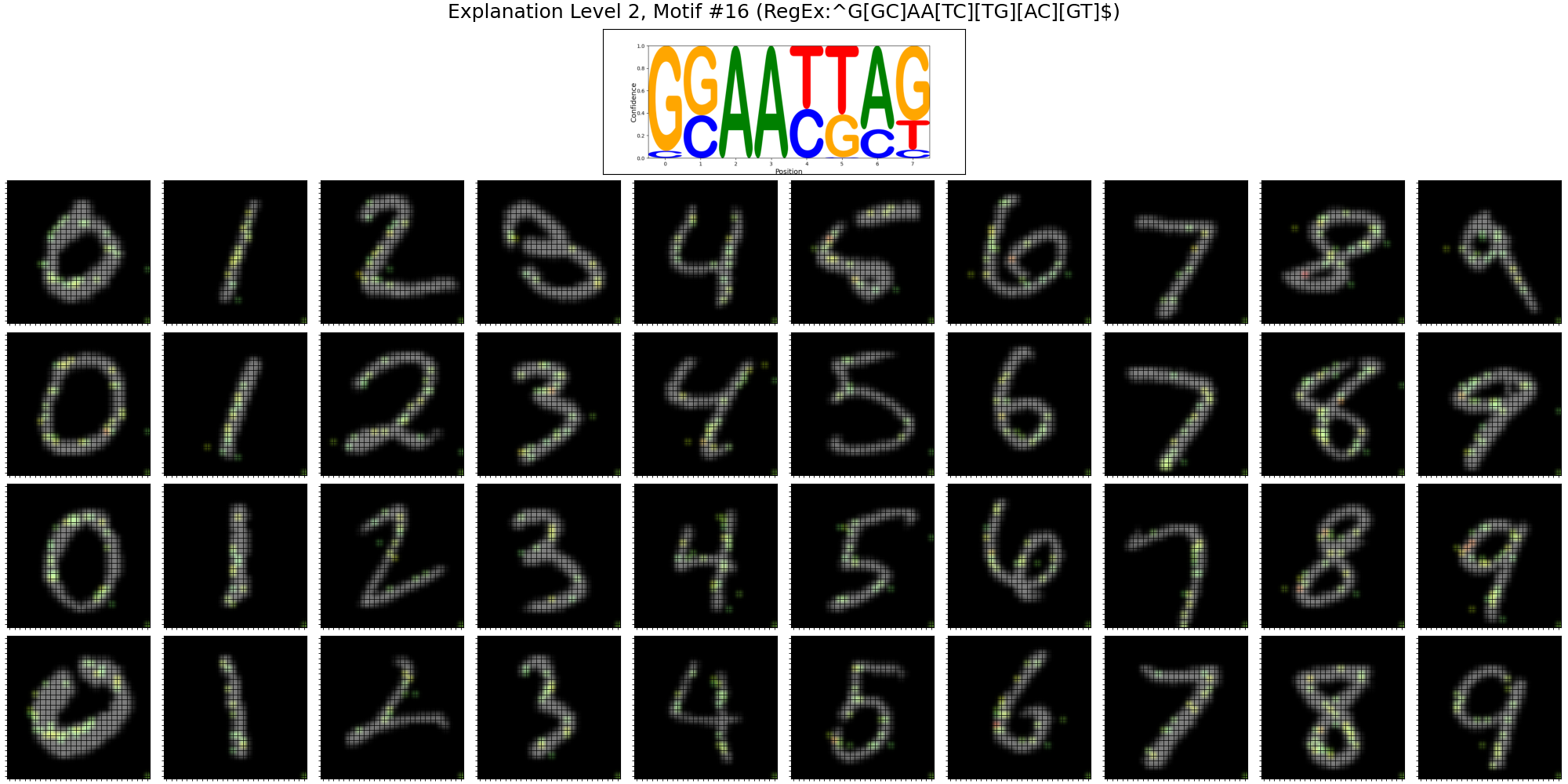}
    \caption{FMotif 15 the most explained effect of level 2 with $FCE=99.65$, caused by a $3 \times 3$ receptive field of FMotifs from level 1. At level 2 each receptive field is mapped to an area $11 \times 11$ of the image input space.}
\end{figure}

\FloatBarrier
\clearpage

\section{Appendix / Detailed explainability metrics}
\label{sec:appF}

\subsection{Motifs of best causality}

\setlength{\tabcolsep}{4pt} 
\begin{table}[!ht]
    \centering
    \caption{Fidelity of cause to effect (FCE) for the best 10 motifs per level of MNIST explanations, using explainer R-ExplaiNet18-16 on the validation set.}
    \begin{tabular}{lcccccccccc}
    \toprule
Level &
1st &
2nd &
3rd &
4th &
5th &
6th &
7th &
8th &
9th &
10th \\
    \midrule
2 & $m^{2}_{16}$ & $m^{2}_{26}$ & $m^{2}_{14}$ & $m^{2}_{79}$ & $m^{2}_{37}$ & $m^{2}_{81}$ & $m^{2}_{63}$ & $m^{2}_{28}$ & $m^{2}_{41}$ & $m^{2}_{85}$ \\
Top-10 FCE & 99.96 & 99.89 & 99.83 & 99.83 & 99.81 & 99.76 & 99.69 & 99.63 & 99.62 & 99.58 \\
~ & ~ & ~ & ~ & ~ & ~ & ~ & ~ & ~ & ~ & ~ \\
3 & $m^{3}_{75}$ & $m^{3}_{14}$ & $m^{3}_{52}$ & $m^{3}_{73}$ & $m^{3}_{35}$ & $m^{3}_{51}$ & $m^{3}_{28}$ & $m^{3}_{76}$ & $m^{3}_{50}$ & $m^{3}_{21}$ \\
Top-10 FCE & 94.93 & 92.34 & 92.21 & 91.85 & 91.62 & 91.02 & 90.33 & 90.10 & 89.91 & 88.22 \\
~ & ~ & ~ & ~ & ~ & ~ & ~ & ~ & ~ & ~ & ~ \\
4 & $m^{4}_{13}$ & $m^{4}_{30}$ & $m^{4}_{46}$ & $m^{4}_{52}$ & $m^{4}_{28}$ & $m^{4}_{39}$ & $m^{4}_{35}$ & $m^{4}_{82}$ & $m^{4}_{60}$ & $m^{4}_{78}$ \\
Top-10 FCE & 99.76 & 99.48 & 98.76 & 98.29 & 98.15 & 97.84 & 97.76 & 97.67 & 97.51 & 96.79 \\
~ & ~ & ~ & ~ & ~ & ~ & ~ & ~ & ~ & ~ & ~ \\
5 & $m^{5}_{73}$ & $m^{5}_{65}$ & $m^{5}_{8}$ & $m^{5}_{1}$ & $m^{5}_{24}$ & $m^{5}_{75}$ & $m^{5}_{64}$ & $m^{5}_{33}$ & $m^{5}_{87}$ & $m^{5}_{61}$ \\
Top-10 FCE & 97.67 & 97.62 & 96.38 & 96.15 & 96.08 & 96.01 & 96.01 & 95.97 & 95.95 & 95.73 \\
~ & ~ & ~ & ~ & ~ & ~ & ~ & ~ & ~ & ~ & ~ \\
6 & $m^{6}_{28}$ & $m^{6}_{40}$ & $m^{6}_{72}$ & $m^{6}_{18}$ & $m^{6}_{39}$ & $m^{6}_{62}$ & $m^{6}_{56}$ & $m^{6}_{2}$ & $m^{6}_{51}$ & $m^{6}_{33}$ \\
Top-10 FCE & 100.00 & 99.85 & 99.75 & 99.74 & 99.73 & 99.69 & 99.68 & 99.66 & 99.66 & 99.65 \\
~ & ~ & ~ & ~ & ~ & ~ & ~ & ~ & ~ & ~ & ~ \\
7 & $m^{7}_{1}$ & $m^{7}_{37}$ & $m^{7}_{57}$ & $m^{7}_{31}$ & $m^{7}_{54}$ & $m^{7}_{15}$ & $m^{7}_{3}$ & $m^{7}_{30}$ & $m^{7}_{48}$ & $m^{7}_{29}$ \\
Top-10 FCE & 99.70 & 99.66 & 97.49 & 97.34 & 97.32 & 96.98 & 96.50 & 96.10 & 96.00 & 95.81 \\
~ & ~ & ~ & ~ & ~ & ~ & ~ & ~ & ~ & ~ & ~ \\
8 & $m^{8}_{38}$ & $m^{8}_{32}$ & $m^{8}_{10}$ & $m^{8}_{20}$ & $m^{8}_{52}$ & $m^{8}_{37}$ & $m^{8}_{8}$ & $m^{8}_{6}$ & $m^{8}_{11}$ & $m^{8}_{41}$ \\
Top-10 FCE & 99.46 & 98.99 & 95.55 & 86.35 & 85.93 & 85.80 & 85.12 & 85.05 & 84.77 & 83.64 \\
~ & ~ & ~ & ~ & ~ & ~ & ~ & ~ & ~ & ~ & ~ \\
    \bottomrule
    \end{tabular}
\end{table}

\FloatBarrier
\clearpage


\newpage
\section*{NeurIPS Paper Checklist}

\begin{enumerate}

\item {\bf Claims}
    \item[] Question: Do the main claims made in the abstract and introduction accurately reflect the paper's contributions and scope?
    \item[] Answer: \answerYes{} 
    \item[] Justification: \textbf{The claims are proven by experiments ,with a publicly available proof-of-concept prototype software implementation, while evidence to support the claims is presented in the sections of the paper.}

\item {\bf Limitations}
    \item[] Question: Does the paper discuss the limitations of the work performed by the authors?
    \item[] Answer: \answerYes{} 
    \item[] Justification: \textbf{Limitations of current work are reported along with future work that is required to overcome them.}

\item {\bf Theory Assumptions and Proofs}
    \item[] Question: For each theoretical result, does the paper provide the full set of assumptions and a complete (and correct) proof?
    \item[] Answer: \answerYes 
    \item[] Justification: \textbf{We present theoretical proof of how the lateral inhibition layer amplifies gradients during gradient descent, that can lead to the attenuation of neuron activations.}

    \item {\bf Experimental Result Reproducibility}
    \item[] Question: Does the paper fully disclose all the information needed to reproduce the main experimental results of the paper to the extent that it affects the main claims and/or conclusions of the paper (regardless of whether the code and data are provided or not)?
    \item[] Answer: \answerYes{} 
    \item[] Justification: \textbf{This work is highly focused on reproducibility. Training epochs are deterministically reproduced in specific hardware+middleware+software setups that are reported in full details, the complete set of hyperparameter configuration for each experiment is provided in the supplementary material in pair with the source code that uses them.}

\item {\bf Open access to data and code}
    \item[] Question: Does the paper provide open access to the data and code, with sufficient instructions to faithfully reproduce the main experimental results, as described in supplemental material?
    \item[] Answer: \answerYes{} 
    \item[] Justification: \textbf{Source code will be openly accessible via GitHub, all datasets used in this work are open, documentation will be uploaded along with the code.}

\item {\bf Experimental Setting/Details}
    \item[] Question: Does the paper specify all the training and test details (e.g., data splits, hyperparameters, how they were chosen, type of optimizer, etc.) necessary to understand the results?
    \item[] Answer: \answerYes{} 
    \item[] Justification: \textbf{The most important  hyperparameters are described in the paper main sections, while other can be found in the Appendices. The full set for every experiment can be found in the supplementary material, where each experiment has a uniquely identifying code that names a hyperparameter configuration file in an easy-to-read .JSON format. }

\item {\bf Experiment Statistical Significance}
    \item[] Question: Does the paper report error bars suitably and correctly defined or other appropriate information about the statistical significance of the experiments?
    \item[] Answer: \answerYes{} 
    \item[] Justification: \textbf{We have trained different architectural setups (layers,features) over 10 random initial conditions and report metrics with confidence intervals.}

\item {\bf Experiments Compute Resources}
    \item[] Question: For each experiment, does the paper provide sufficient information on the computer resources (type of compute workers, memory, time of execution) needed to reproduce the experiments?
    \item[] Answer: \answerYes{} 
    \item[] Justification: \textbf{The compute resource requirements are mentioned. All experiments run on a single worker, where secs/epoch for each experiment is recorded in training logs that will be in the supplementary material. Indicative measurements of elapsed time for training are presented for some experiments in the Appendices. }
    
\item {\bf Code Of Ethics}
    \item[] Question: Does the research conducted in the paper conform, in every respect, with the NeurIPS Code of Ethics \url{https://neurips.cc/public/EthicsGuidelines}?
    \item[] Answer: \answerYes{}
    \item[] Justification: \textbf{Using public datasets ensures compliance to worldwide privacy laws. Since our models are not generative no provisions are needed for adding indications or watermarking on their output.}

\item {\bf Broader Impacts}
    \item[] Question: Does the paper discuss both potential positive societal impacts and negative societal impacts of the work performed?
    \item[] Answer: \answerYes{} 
    \item[] Justification: \textbf{The specific field of research that our work belongs to, targets to bring positive societal impacts, mainly increasing trustworthiness, ensuring compliance with law and mitigating model bias which are mentioned in the paper.} 
    
\item {\bf Safeguards}
    \item[] Question: Does the paper describe safeguards that have been put in place for responsible release of data or models that have a high risk for misuse (e.g., pretrained language models, image generators, or scraped datasets)?
    \item[] Answer: \answerNA{} 
    \item[] Justification: 

\item {\bf Licenses for existing assets}
    \item[] Question: Are the creators or original owners of assets (e.g., code, data, models), used in the paper, properly credited and are the license and terms of use explicitly mentioned and properly respected?
    \item[] Answer: \answerYes{} 
    \item[] Justification: \textbf{Citations, preserving license comments/files in the source code, attribution with footnote links to dataset homepages, attribution with short description of work.}

\item {\bf New Assets}
    \item[] Question: Are new assets introduced in the paper well documented and is the documentation provided alongside the assets?
    \item[] Answer: \answerYes{} 
    \item[] Justification: \textbf{New code/model assets are introduced in the paper, detailed documentation will be provided as part of a  GitHub repository or/and project homepage.}

\item {\bf Crowdsourcing and Research with Human Subjects}
    \item[] Question: For crowdsourcing experiments and research with human subjects, does the paper include the full text of instructions given to participants and screenshots, if applicable, as well as details about compensation (if any)? 
    \item[] Answer: \answerNA{}
    \item[] Justification:

\item {\bf Institutional Review Board (IRB) Approvals or Equivalent for Research with Human Subjects}
    \item[] Question: Does the paper describe potential risks incurred by study participants, whether such risks were disclosed to the subjects, and whether Institutional Review Board (IRB) approvals (or an equivalent approval/review based on the requirements of your country or institution) were obtained?
    \item[] Answer: \answerNA{} 
    \item[] Justification: 

\end{enumerate}

\end{document}